\documentclass{article}

\usepackage{microtype}
\usepackage{graphicx}
\usepackage{subfigure}
\usepackage{booktabs} 
\usepackage{placeins}
\usepackage{multirow}
\usepackage{natbib}
\usepackage{amsmath, amsfonts}
\usepackage{amsthm}
\usepackage{thmtools}
\usepackage{thm-restate}
\usepackage{algorithm}
\usepackage{algorithmic}
\usepackage[margin=1in]{geometry}

\usepackage[colorlinks=true, urlcolor=magenta, citecolor=blue, pagebackref=true]{hyperref} 
\renewcommand*\backref[1]{\ifx#1\relax \else (Cited on #1) \fi}
 
\theoremstyle{plain}
\newtheorem{theorem}{Theorem}[section]

\newtheorem{lemma}[theorem]{Lemma}
\newtheorem{corollary}[theorem]{Corollary}
\theoremstyle{definition}
\newtheorem{definition}[theorem]{Definition}
\newtheorem{problem}[theorem]{Problem}

\theoremstyle{remark}
\newtheorem{remark}[theorem]{Remark}

\usepackage[textsize=tiny]{todonotes}

\DeclareMathOperator*{\argmax}{arg\,max}
\DeclareMathOperator*{\argmin}{arg\,min}

\newcommand{\R}{\mathbb{R}}

\newcommand{\txt}[1]
{\textnormal{#1}}

\newcommand{\E}[2]
{\mathbb{E}_{#1}[{#2}]}

\newcommand{\EE}[2]
{\mathbb{E}_{#1}\left[{#2}\right]}

\newcommand{\mc}[1]
{\mathcal{#1}}

\newcommand{\algname}{\txt{Nonlinear Low-Rank Approximation }}

\newcommand{\algnameshort}{\txt{NLRA }}

\let\oldeqref\eqref
\renewcommand\eqref[1]{Eq.~\oldeqref{#1}}

\newcommand{\eps}{\epsilon}

\usepackage{dsfont}
\usepackage{amssymb,amsmath}

\newcommand\NLRA{\operatorname{\textbf{NLRA}}} 

\usepackage{hyperref}

\title{Nonlinear Initialization Methods for Low-Rank Neural Networks}

\author{%
  \hspace{-1em}Kiran Vodrahalli\thanks{Columbia University. Work performed at Google. Email: \texttt{kiran.vodrahalli@columbia.edu}}\hspace{-1em}
  \and Rakesh Shivanna\thanks{Google Research, Brain Team. Email: \texttt{rakeshshivanna@google.com}}\hspace{-1em}
  \and Maheswaran Sathiamoorthy\thanks{Google Research, Brain Team. Email: \texttt{nlogn@google.com}}\hspace{-1em}
  \and Sagar Jain\thanks{Google Research, Brain Team. Email: \texttt{sagarj@google.com}}\hspace{-1em}
  \and Ed H. Chi\thanks{Google Research, Brain Team. Email: \texttt{edchi@google.com}}\hspace{-1em}
}
\date{}

\begin{document}

\maketitle


\begin{abstract}

We propose a novel low-rank initialization framework for training low-rank deep neural networks -- networks where the weight parameters are re-parameterized by products of two low-rank matrices. The most successful prior existing approach, spectral initialization, draws a sample from the initialization distribution for the full-rank setting and then optimally approximates the full-rank initialization parameters in the Frobenius norm with a pair of low-rank initialization matrices via singular value decomposition.  Our method is inspired by the insight that approximating the $\textit{function}$ corresponding to each layer is more important than approximating the parameter values. We provably demonstrate that there is a significant gap between these two approaches for ReLU networks, particularly as the desired rank of the approximating weights decreases, or as the dimension of the inputs to the layer increases (the latter point holds when the network's width is superlinear in dimension). Along the way, we provide the first provably efficient algorithm for solving the ReLU low-rank approximation problem for fixed parameter rank $r$ -- previously, it was unknown that the problem was computationally tractable to solve even for rank $1$. We also provide a practical algorithm to solve this problem which is no more expensive than the existing spectral initialization approach, and validate our theory by training ResNet and EfficientNet models \citep{he2016deepresidual, tan2019efficientnet} on ImageNet \citep{ILSVRC15}.
\end{abstract}

\section{Introduction} \label{sec:intro}
\looseness=-1
Training deep networks is a canonical task in modern machine learning. Training and serving deep networks with very large parameter counts efficiently is of paramount importance in recent years, since significant performance gains on a variety of tasks are possible simply by scaling up parameter counts \citep{rae2021scaling, shin2021scaling, henighan2020scaling, gordon2021data, sharma2020neural, kaplan2020scaling}. Further theoretical evidence suggests that requiring the resulting learned networks to be smooth functions (and therefore, in some sense, robust to perturbations) entails even larger parameter counts \citep{bubeck2021universal}. 

However, there are significant computational difficulties with both training and deploying such large models. Most existing works focus on efficiently deploying trained deep networks and uses a variety of approaches including sparsifying neural network weights \citep{reed1993pruning, blalock2020state, lecun1990optimal, hassibi1993second, mozer1989skeletonization, dong2017learning, han2016deep, lewis2021base, yang2019deephoyer, su2020sanitychecking, frankle2020pruning, fischer2021plant}, model distillation \citep{heo2019knowledge, hinton2015distilling}, low-rank post-processing \citep{anonymous2022language, idelbayev2020low, tukan2021no}, and mixtures thereof \citep{chen2021scatterbrain}. Unfortunately, these methods do not address the question of efficient training. 

\subsection{Training-Time Efficient Deep Networks}
Another general approach to improving speed and memory at both train \textit{and} inference time is to replace each weight matrix $W \in \mathbb{R}^{d \times m}$ with a computationally efficient representation which improves the training speed of standard gradient-based optimization methods, while ideally not losing out on the representation capacity too much. A variety of papers have taken this approach with varying techniques including fast Fourier transform-inspired sparse matrix decompositions, low-rank factorizations, orthogonal basis kernel approximations, and sketching-based approaches \citep{dao2020kaleidoscope, khodak2021initialization, vahid2020butterfly, wang2021pufferfish, dusenberry2020efficient, romero2014fitnets, zhang2015accelerating, choromanski2020rethinking, chen2021mongoose, panigrahy2021sketch}. In the sparse deep network literature, there also exist methods to perform pruning before or during training (to also get efficiency gains during training) \citep{wang2019picking, mostafa2019parameter, he2021learning, su2020sanitychecking, frankle2020pruning, mussay2020dataindependent}.

\subsection{Low-Rank Factorized Networks}
\looseness=-1
Given the broad spectrum of proposed methodologies, it may be unclear to practitioners which families of methods are most practical to use. Two recently popular approaches are unstructured sparse pruning and low-rank factorization. Low-rank methods do not require specialized hardware to convert smaller parameter counts into compute savings, unlike the sparse methods \citep{paszke2019pytorch, nvidia2020, mishra2021accelerating}. 

For a fully-connected layer, the basic idea of a low-rank layer (or a \textit{factored layer}) is to parameterize the network weights $W\in\R^{d\times m}$ with a low-rank matrix product $UV^{\top}$, where $U \in \mathbb{R}^{d \times r}, V \in \mathbb{R}^{m \times r}$ and $r$ is the (user-selected) rank. Simple generalizations exist for other standard layers including convolution and attention layers \citep{khodak2021initialization}. Low-rank deep networks reduce parameter counts (thus saving memory) as well as the number of ops required for matrix-vector multiplication: $(d + m)\cdot r$ vs. $d\cdot m$. 

\citet{khodak2021initialization} demonstrate that if one pays attention to proper initialization and regularization, low-rank methods outperform sparse pruning approaches in many domains, contrary to existing beliefs that sparse methods outperform low-rank methods in parameter count savings.
In particular, a low-rank initialization scheme called \textbf{spectral initialization} is crucial to achieve better performance -- initialization schemes are in general quite important for achieving good performance in neural network training \citep{bachlechner2020rezero, choromanski2018initialization, dauphin2019metainit, hu2020provable, huang2020improving, mishkin2015all, pennington2017resurrecting, xiao2018dynamical, zhang2021initialization}.
Spectral initialization samples a full-rank matrix $W \in \R^{d \times m}$ from a known init distribution, factorizes $W$ as $A\Sigma^{1/2}, \Sigma^{1/2}B^{\top}$ via singular value decomposition (SVD), and initializes $U$ and $V^\top$ with these factors.

However, there are no explanations for why this approach yields improved performance beyond the zeroth-order justification that it well-approximates the parameters of the full-rank weights at initialization. 
Thus, it is a natural next step to develop better theoretical understanding of the properties of the low-rank network learning problem, with the hope that it will aid us in finding improved methods for training low-rank versions of deep networks and in uncovering the principles of learning low-rank network approximations.

\subsection{Low-Rank Approximation Theory} 

At the heart of the approach we take (see Problem~\ref{prob:learning_prob_statement} in Section~\ref{sec:properties_nonlinear_lowrank}) is the idea that we want to make an alternate (nonlinear) low-rank approximation to a matrix $W$. Low-rank approximation has been long studied in the theoretical computer science literature (see \citep{woodruff2014sketching, mahoney2011randomized} for thorough surveys of the topic). One particularly related line of work is the masked low-rank approximation literature \citep{musco2019simple}. The basic idea of masked low-rank approximation is that we want to find a low-rank $Y$ that minimizes $\left\|M \circ \left(W - Y\right)\right\|_F^2$, where $M$ is a mask applied as an elementwise product (the Hadamard product). This problem captures many different problems studied in the literature under various structural assumptions on $M$ (for instance, the case where $M$ is a real-valued non-negative is known as weighted low-rank approximation, and was studied by \citet{razenshteyn2016weighted}). \citet{musco2019simple} study the case where $M$ is a binary mask and provide bicriteria approximation guarantees since the problem is hard in general. It is interesting to consider the connection between the binary masked low-rank approximation problem of \citet{musco2019simple} and our problem, where we apply ReLU to \textit{Gaussian samples} and mask only the \textit{output} --- interestingly, in our case, the problem is solvable efficiently. It would be interesting to establish further connections between our problem setting and other low-rank approximation settings, perhaps by adopting our setup from Problem~\ref{prob:learning_prob_statement} but possibly changing the input distribution. It seems plausible that for some choices of input distribution, one could make the problem computationally hard. For more background on the low-rank approximation literature, see the discussion in \citet{musco2019simple}.

\subsection{Contributions} 
We study the problem of learning low-rank neural networks in this paper and ask:
\begin{enumerate}
    \item \textit{Are there better initialization schemes for low-rank deep neural networks that improve post-training generalization error?}
    \item \textit{What factors govern the theory of choosing a low-rank initialization method?}
\end{enumerate}

Our contributions are as follows:

\begin{enumerate}
	\item The identification of the \textbf{function approximation at initialization} framework (Definitions~\ref{def:function_approx_at_init}, ~\ref{def:layerwise_func_approx_init}) and the simpler (and parallelized) \algname objective ($\NLRA$) (Problem~\ref{prob:learning_prob_statement}) for initializing low-rank networks (and it also applies to other structured network approximation schemes);
	\item A provably efficient algorithm (Algorithm~\ref{alg:relu_svd}) to solve the resulting ReLU low-rank approximation problem for fixed rank $r$ (to the best of our knowledge it was unknown that it was computationally tractable to solve even for rank $1$); 
	\item Practical algorithms (Algorithms~\ref{alg:lfai_gradient}, ~\ref{alg:lfai_ws_gradient}) and corresponding empirical results which validate that our layerwise low-rank initialization scheme works as an efficient drop-in replacement for the commonly used spectral initialization with improved performance (on average $0.3\%$ top-$1$ accuracy gain on ImageNet \citep{ILSVRC15}, and as much as up to $1.2\%$ accuracy improvement);
	\item Empirical observations that taking nonlinearities of layers into account for the initialization scheme improves accuracy for lower-rank layers and larger input dimension and width;
	\item Theoretical proof that in settings used in practice, the sub-optimality of the spectral initialization with respect to the $\NLRA$ objective grows with decreasing rank, and also input dimension (as long as the layer width is super-linear in the input dimension);
	\item Empirical confirmation that optimization of the $\NLRA$ objective at initialization positively correlates with decreasing post-training test error.
\end{enumerate}

\subsection{Paper Outline}
In Section~\ref{sec:approx_at_init}, we introduce our novel initialization framework (Definition~\ref{def:function_approx_at_init}) and present a special case (Definition~\ref{def:layerwise_func_approx_init}) which corresponds to the $\algname$ problem ($\NLRA$). We provide practical heuristic algorithms (Algorithms~\ref{alg:lfai_gradient}, ~\ref{alg:lfai_ws_gradient}) to solve this problem, which we use in experiments. In Section~\ref{sec:properties_nonlinear_lowrank}, we lower bound the gap in performance between the spectral solution and the optimal solution to the $\NLRA$ problem, and characterize the problem parameters for which this gap is large (Theorem~\ref{thm:subopt_lowerbound} and Corollary~\ref{cor:spherical_weights_dim_increase_rank_decrease}).  We also give an (impractical) polynomial time algorithm (Algorithm~\ref{alg:relu_svd}) to solve this problem when the nonlinearity is the ReLU and the target rank $r$ is constant, and prove its correctness and efficiency (Theorem~\ref{thm:relu_svd}).
In Section~\ref{sec:experiments}, we analyze our experimental results, and we conclude in Section~\ref{sec:conclusion}. 
In Appendix~\ref{appendix:proofs}, we prove many of the theorems stated in the main body of the paper. In Appendix~\ref{appendix:beyond_LRA}, we outline methodology based on function approximation at initialization which extends beyond low-rank approximation to any layer which approximates a different layer (for instance, to any approximate version of a Transformer \citep{choromanski2020rethinking}).

\section{Function Approximation at Initialization} \label{sec:approx_at_init}

In this section, we propose a general framework for initializing low-rank deep networks given a full-rank initialization scheme. 
We also describe a special case of our framework which is provably efficient to implement when the nonlinearity is ReLU and the target rank $r$ is a constant.

\subsection{Function Approximation at Init Framework}

The key idea behind our approach is to mimic the full-rank initialization distribution as closely as possible. \citet{khodak2021initialization} follows this principle to argue for the spectral initialization approach to initialize the low-rank weights: the idea is to match the weight matrices in Frobenius norm as closely as possible using SVD. More precisely:

\begin{definition}[Spectral Initialization]\label{def:spectral_init}
Given a full-rank weight $W \in \mathbb{R}^{d \times m}$ initialized according to distribution $\mathcal{D}$, spectral initialization with rank $r$ is the following procedure:

\begin{enumerate}
    \item Sample $W \sim \mathcal{D}$.
    \item Factorize $U\Sigma V^{\top} = W$ via singular value decomposition, where $U \in \mathbb{R}^{d \times r}, \Sigma \in \mathbb{R}^{r \times r}, V \in \mathbb{R}^{m \times r}$.
    \item Output factor initialization $\hat{U} = U\Sigma^{1/2}$, $\hat{V} = V\Sigma^{1/2}$.
\end{enumerate}
\end{definition}

However, it is not clear that this metric for matching low-rank to full-rank functions accurately captures what is important in the approximation: Ultimately, we are initializing highly nonlinear functions, and various terms in the weights may be less important than others (in a post-training setting, \citet{anonymous2022language} demonstrates the efficacy of taking nonlinearities into account). Thus, we consider a function-approximation viewpoint rather than a weight-approximation viewpoint at initialization:

\begin{definition}[Function Approximation at Initialization] \label{def:function_approx_at_init}
Given a full-rank weight distribution $\mathcal{D}$, we find a low-rank weight matrix of rank $r$ for initialization as follows:

\begin{enumerate}
    \item Sample $W = [W_1, \cdots, W_k] \sim \mathcal{D}$ with $W_i \in \mathbb{R}^{d_i \times m_i}$.
    \item Solve 
    $\hat{U}, \hat{V} = \argmin_{U, V} \mathbb{E}_{x \sim \mathcal{N}(0, I)}\big[\big(f_{U, V}(x) - f_W(x)\big)^2\big]
    $,
    where $f$ denotes the deep network class we consider, and $\hat{U} = [\hat{U}_1, \cdots, \hat{U}_k]$ and $\hat{V} = [\hat{V}_1, \cdots, \hat{V}_k]$, where $\hat{U}_i \in \mathbb{R}^{d_i \times r}$ and $\hat{V}_i \in \mathbb{R}^{m_i \times r}$.
    \item Use $\hat{U}, \hat{V}$ as the initialization for the low-rank weights.
\end{enumerate}
\end{definition}

Thus, we attempt to match the function values of the networks, rather than simply the weights, making our approach \textit{nonlinearity-aware}. Here we have chosen to measure the similarity of the network outputs over a standard Gaussian input distribution; however, this aspect can easily be modified to be samples over a particular distribution of interest.

Since this approach is essentially function approximation of the initialization network (rather than the trained network, as in other work), we refer to our general approach as \textit{function approximation at initialization}.
To implement this approach, one can optimize the low-rank parameters with a gradient-based method with some automatic differentiation framework like Tensorflow \citep{abadi2015tensorflow}.
We now proceed to outline a simplification to this general approach which is more tractable and easier to use in practice.




\subsection{Layerwise Function Approximation at Initialization} \label{subsec:layerwise_func_approx_init}


 
To make the approximation problem more tractable to solve in practice, we propose a simplification to the general approach: keep the relevance of the non-linearity, but instead approximate each layer separately rather than the entire network. This approach has the benefits of a) being a simpler problem to solve, and b) being embarrassingly parallel to distribute. 
Thus, we can obtain a significant speedup in the initialization method compared to the full function approximation at initialization approach.

We propose an empirical sample-based stochastic optimization approach for solving the problem using gradient methods:
\begin{definition}[Layerwise Function Approximation at Initialization (LFAI)] \label{def:layerwise_func_approx_init}
Given a deep neural network $f$, define the function corresponding to the $i^{\txt{th}}$ layer with weights $W_i$ as $f_{W_i}$. Then, given a full-rank weight distribution $\mathcal{D}$, we find a low-rank weight matrix of rank $r$ for initialization as follows:
\begin{enumerate}
    \item Sample $W_i \sim \mathcal{D}$;
    \item Solve  
    $
    \hat{U}_i, \hat{V}_i = \argmin_{U_i, V_i} \mathbb{E}_{x \sim \mathcal{N}(0, I)}\big[ \big(f_{U_i, V_i}(x) - $ $ f_{W_i}(x)\big)^2\big]
    $
    for all layers $i$ in parallel;
    \item Use $\hat{U}_i, \hat{V}_i$ as the low-rank initialization for layer $i$.
\end{enumerate}
\end{definition}


We can optimize the parameters directly using some gradient-based method over Gaussian\footnote{See Remark~\ref{rem:choice_of_input} for a discussion of other input distributions.} samples. In this case, we are essentially throwing out information about $W_i$, since we only access information about $W_i$ via samples which are fed into the gradient-based method. This algorithm is run in parallel across all layers of the network. We present the algorithm at a single layer in Algorithm~\ref{alg:lfai_gradient}.

\begin{algorithm}
\caption{LFAI-Gradient}\label{alg:lfai_gradient}
\begin{algorithmic}[1]
\REQUIRE $r < d < m$, sample access to full-rank init distribution $\mathcal{D}$ over $\R^{d \times m}$, iterative gradient method $\mathcal{A}$, number of samples $N$
\STATE Sample $W \sim \mathcal{D}$.
\STATE Sample $\{x_k\}_{k = 1}^N$ i.i.d. from $\mathcal{N}(0, I_{d \times d})$.
\STATE Run $\mathcal{A}$ using gradient $\nabla_{U, V} \frac{1}{N}\sum_{k = 1}^N \big[ \big(f_{U, V}(x_k) - f_{W}(x_k)\big)^2\big]$ until convergence.
\STATE \textbf{Return:} $\hat{U} \in \R^{d \times r}$, $\hat{V} \in \R^{m \times r}$.
\end{algorithmic}
\end{algorithm}

To further improve efficiency, we can feed the initialization algorithm some prior information about $W$ by initializing the low-rank weights with spectral initialization (Definition~\ref{def:spectral_init}) -- we call this step the ``spectral warm start,'' and observe that it helps with learning empirically (see Section~\ref{sec:experiments}). We can think of this step as reducing the sample complexity required for optimizing only from samples $(x_k, f(x_k))$. The algorithm is presented in Algorithm~\ref{alg:lfai_ws_gradient}.

\begin{algorithm}
\caption{LFAI-WS-Gradient}\label{alg:lfai_ws_gradient}
\begin{algorithmic}[1]
\REQUIRE $r < d < m$, sample access to full-rank init distribution $\mathcal{D}$ over $\R^{d \times m}$, iterative gradient method $\mathcal{A}$, number of samples $N$
\STATE Sample $W \sim \mathcal{D}$.
\STATE Sample $\{x_k\}_{k = 1}^N$ i.i.d. from $\mathcal{N}(0, I_{d \times d})$.
\STATE Compute rank-$r$ SVD $W = U_r\Sigma_r V_r^{\top}$.
\STATE Initialize $U_0 := U_r\Sigma_r^{1/2}$; $V_0 := V_r\Sigma_r^{1/2}$.
\STATE Run $\mathcal{A}$ using gradient $\nabla_{U, V} \frac{1}{N}\sum_{k = 1}^N \big[ \big(f_{U, V}(x_k) - f_{W}(x_k)\big)^2\big]$ until convergence.
\STATE \textbf{Return:} $\hat{U} \in \R^{d \times r}$, $\hat{V} \in \R^{m \times r}$.
\end{algorithmic}
\end{algorithm}

\begin{table*}[t]
\centering
\caption{The reported metrics are average top-$1$ accuracy as a percent on EfficientNet models of increasing width and depth, trained with Weight Decay regularization. We report standard error over three samples up to one standard deviation. The EfficientNet model variants we consider have width and depth scale parameters set to be $(1, 1); (1.2, 1.4); (2.0, 3.1); (3.0, 3.2) $, corresponding to the b$0$, b$3$, b$7$, and b$9$ variants respectively. We see the empirical effect described in Section~\ref{sec:properties_nonlinear_lowrank} -- our methods (LFAI-Adam and LFAI-WS-Adam) have more significant improvements for smaller rank scales and larger width models.}
\label{table:effnet_results}
\vskip 0.15in
\begin{tabular}{|c|c|c|c|c|c|}
\hline
 Rank Scale & Method & EfficientNet-b9 & EfficientNet-b7 & EfficientNet-b3 & EfficientNet-b0 \\ \hline
 \multirow{4}{*}{0.05} & Baseline & $66.36 \pm 0.23$ & $58.61 \pm 0.45$ & $37.34 \pm 0.29$ & $31.22 \pm 0.62$\\ \cline{2-6}
 & Spectral &  $66.64 \pm 0.23$ & $58.84 \pm 0.38$ & $\mathbf{38.3 \pm 0.09}$ & $30.96 \pm 0.32$\\ \cline{2-6}
 & \textbf{LFAI-Adam} & $65.84 \pm 0.28$ & $59.00 \pm 0.18$ & $\mathbf{38.22 \pm 0.28}$ & $\mathbf{32.39 \pm 0.26}$\\ \cline{2-6}
 & \textbf{LFAI-WS-Adam} & $\mathbf{66.90 \pm 0.11}$ & $\mathbf{59.63 \pm 0.37}$ & $38.04 \pm 0.03$ & $30.8 \pm 0.19$ \\ \hline
 \multirow{4}{*}{0.10} & Baseline & $71.67 \pm 0.19$ &  $68.89 \pm 0.07$ & $\mathbf{57.06 \pm 0.24}$ & $\mathbf{48.24 \pm 0.13}$\\ \cline{2-6}
 & Spectral & $72.07 \pm 0.15$ & $68.83 \pm 0.16$ & $56.66 \pm 0.35$ & $47.79 \pm 0.17$\\ \cline{2-6}
 & LFAI-Adam & $71.02 \pm 0.32$ & $68.52 \pm 0.04$ & $\mathbf{56.96 \pm 0.19}$ & $47.42 \pm 0.30$\\ \cline{2-6}
 & \textbf{LFAI-WS-Adam} & $\mathbf{72.28 \pm 0.04}$ & $\mathbf{69.09 \pm 0.16}$ & $56.76 \pm 0.11$ & $47.84 \pm 0.24$\\ \hline
 \multirow{4}{*}{0.15} & Baseline & $73.20 \pm 0.05$ &  $71.67 \pm 0.19$ & $62.56 \pm 0.07$ & $54.99 \pm 0.05$\\ \cline{2-6}
 & Spectral & $\mathbf{73.55 \pm 0.15}$ & $71.55 \pm 0.20$ & $63.02 \pm 0.07$ & $54.97 \pm 0.10$ \\ \cline{2-6}
 & LFAI-Adam & $72.70 \pm 0.18$ & $71.25 \pm 0.12$ & $62.86 \pm 0.12$ & $54.85 \pm 0.09$\\ \cline{2-6}
 & \textbf{LFAI-WS-Adam} & $73.31 \pm 0.04$ & $\mathbf{71.74 \pm 0.02}$ & $\mathbf{63.55 \pm 0.16}$ & $\mathbf{55.19 \pm 0.10}$\\ \hline
  \multirow{4}{*}{0.20} & Baseline & $73.84 \pm 0.04$ &  $72.16 \pm 0.27$ & $66.05 \pm 0.23$ & $\mathbf{59.78 \pm 0.17}$\\ \cline{2-6}
 & Spectral & $\mathbf{74.14 \pm 0.19}$ & $72.45 \pm 0.17$ & $65.96 \pm 0.29$& $59.42 \pm 0.32$ \\\cline{2-6}
 & LFAI-Adam & $73.91 \pm 0.09$ & $72.32 \pm 0.18$ & $65.93 \pm 0.06$ &$59.20 \pm 0.11$\\ \cline{2-6}
 & \textbf{LFAI-WS-Adam} & $\mathbf{74.01 \pm 0.11}$ & $\mathbf{72.72 \pm 0.03}$ & $\mathbf{66.32 \pm 0.02}$ &$59.38 \pm 0.18$\\ \hline
 1.0 & Full-Rank & $79.62 \pm 0.05$ & $78.70 \pm 0.02$ & $76.63 \pm 0.19$ & $74.50 \pm 0.13$\\ \hline
\end{tabular}
\end{table*}

\section{Nonlinear Low-Rank Approximation} \label{sec:properties_nonlinear_lowrank}

In this section, we demonstrate settings in which we can expect the results of using layer-wise function approximation at initialization for the ReLU activation to be significantly different from using spectral initialization, and we also resolve an open theoretical question on the computational tractability of low-rank ReLU approximation. Throughout this section, we will refer to $\sigma_{\txt{ReLU}}$ as $\sigma$ for simplicity. First we instantiate the layer-wise function approximation objective for fully-connected layers:

\begin{problem}[Nonlinear Low-Rank Approximation (NLRA)]\label{prob:learning_prob_statement}
Consider the objective
\begin{align}
    \mathcal{R}(Y) 
    := \mathbb{E}_{x \sim \mathcal{N}(0, I)}\left[\left\|\sigma_{\txt{ReLU}}(x^\top Y) - \sigma_{\txt{ReLU}}(x^\top W)\right\|_2^2\right] \label{eq:population_objective}
\end{align}
where $\sigma_{\txt{ReLU}}(x) = \max(0, x)$ is the ReLU activation, $W, Y \in \mathbb{R}^{d \times m}$, and $W$ are fixed ground-truth full-rank weights.
Let $\txt{opt} := \mathcal{R}\left(Y^*\right)$, where $Y^*$ is the argmin over matrices of rank $r$. Our goal is to give a computationally efficient algorithm for outputting a rank $r$ matrix $\hat{Y}$ such that
$
\mathcal{R}(\hat{Y})  < \txt{opt} + \epsilon
$.
\end{problem}

We will derive some structural properties of this objective which will allow us to a) give a fixed-parameter tractable polynomial time algorithm for solving the problem when the rank $r$ is constant, and b) lower bound the gap between the quality of the spectral solution (linear approximation) and the quality of the (optimal) nonlinear low-rank approximation with respect to this nonlinear error measure.

Using our analysis, we uncover some conditions on the full-rank matrix which yield a more significant gap. In particular, as the rank gets smaller or as the dimension of the inputs increases (while the width is super-linear in the input dimension), the gap between the initialization methods blows up with dimension (Theorem~\ref{thm:subopt_lowerbound} and Corollaries~\ref{cor:small_rho_larger_gap}, \ref{cor:dim_increase_rank_decrease}). It is also the case that full-rank matrices $W \in \R^{d \times m}$ with more approximately orthogonal columns yields a larger gap. As a technical tool, we prove a characterization of the nonlinear function approximation problem (Theorem~\ref{thm:relu_svd}), which applies specifically to one-hidden-layer ReLU networks (rather than easily invertible activations). We then exploit the properties of this characterization to prove Theorem~\ref{thm:subopt_lowerbound}. This characterization also yields an efficient algorithm for solving $\NLRA$ in the case when the target rank $r$ is a fixed constant. 
We defer all proofs and full theorem statements to Section~\ref{subsec:proofs_sec3} of the Appendix. Despite the restriction to ReLU, we believe our results to be characteristic for other activation functions, as we empirically demonstrate for the Swish activation in Section~\ref{sec:experiments}.

These results only apply to the difference between SVD (the spectral solution) and the optimal nonlinear low-rank approximation. While this result is useful for understanding what properties of the full-rank matrix $W$ govern the extreme cases where the two initialization approaches are very similar or very different, these initialization results do not directly prove anything about downstream generalization error, except in the setting where the true optimal weights are close to the initialization distribution.
Nevertheless, we empirically demonstrate the connection between good nonlinear low-rank approximation and improved downstream generalization error for deep low-rank models in Section~\ref{sec:experiments}.

\subsection{An Efficient $\NLRA$ Algorithm for the ReLU Activation}

We prove a lower bound on the gap between the output of spectral initialization and our \algnameshort method for one-hidden-layer ReLU network. To achieve this bound, we further develop theory characterizing the optimal low-rank matrix for Problem~\ref{prob:learning_prob_statement} for the ReLU activation.
We begin with a useful definition -- we will use this function in our analysis.

\begin{definition}[ReLU Kernel: $1^{st}$-Order Arc-Cosine Kernel]\label{def:arccos_kernel}
The first-order arc-cosine kernel \citep{cho2009kernel} is defined by 
$
k(x, y) := \|x\|_2 \|y\|_2 \cdot \sqrt{h}(\rho_{xy})
$,
where
$
\rho_{xy} := \frac{x^{\top}y}{\|x\|_2\|y\|_2}
$
and
$
\sqrt{h}(\rho_{xy}) = \big(\sqrt{1 - \rho_{xy}^2} + (\pi - \cos^{-1}(\rho_{xy}))\rho_{xy}\big)\slash{\pi}
$.
\end{definition}

Now we present an alternative algorithm to solve the \algnameshort problem for the ReLU activation. This algorithm runs in polynomial time when the target rank $r$ is constant; however, its utility comes from the characterization of the optimal solution to the ReLU \algnameshort problem -- in particular, it reveals more structure of the ReLU kernel which allows us to easily lower bound the sub-optimality of standard Frobenius low-rank approximation computed using SVD. 
\begin{theorem}[ReLU SVD]\label{thm:relu_svd}
Consider the goal of finding the optimal rank-$r$ solution to the objective $\mc{R}(Y)$ in Problem~\ref{prob:learning_prob_statement} with known $W \in \R^{d \times m}$. 
Suppose we solve the following problem:
\[ 
\Lambda^* = \argmax_{\substack{\Lambda \in \{0, 1\}^{d};~\|\Lambda\|_1 = r}} \quad \sum_{i = 1}^m \|W_i\|_2^2\cdot h\left(\frac{\left\|\Sigma \txt{ diag}\left(\Lambda\right)V_i\right\|_2}{\|\Sigma V_i\|_2}\right)
\]
where $W = U\Sigma V^{\top}$ with $U \in \mathbb{R}^{d \times d}, \Sigma \in \mathbb{R}^{d \times d}$ is diagonal, and $V \in \mathbb{R}^{m \times d}$, with $U^{\top}U = I_{d \times d}, V^{\top}V = I_{d \times d}$ and $V_i \in \mathbb{R}^d$ is the $i^{th}$ column of $V^{\top}$, and where $h$ is defined in Definition~\ref{def:hfunc}.
Then, the optimal rank-$r$ matrix $Y^* \in \mathbb{R}^{d \times m}$ minimizing the initial objective ~\eqref{eq:population_objective} is given by 
$
Y^* = U \txt{ gnorm}\left(\txt{ diag}\left(\Lambda^*\right)\Sigma V^{\top}\right),
$
where $
\txt{gnorm}(A)_i := A_i \cdot \frac{\sqrt{h}(\|A_i\|_2)}{\|A_i\|_2}
$
for the $i^\txt{th}$ column of matrix $A$.
\end{theorem}
\begin{proof}
See the appendix.
\end{proof}

Theorem~\ref{thm:relu_svd} directly yields a fixed-parameter tractable algorithm for solving ReLU $\NLRA$:

\begin{algorithm}
\caption{ReLU SVD}\label{alg:relu_svd}
\begin{algorithmic}[1]
\REQUIRE $W \in \mathbb{R}^{d \times m}$, $r < d < m$
\STATE Compute SVD $W = U\Sigma V^{\top}$.
\STATE Compute $\sum_{i = 1}^m \|W_i\|_2^2\cdot h\left(\frac{\left\|\Sigma \txt{ diag}\left(\Lambda\right)V_i\right\|_2}{\|\Sigma V_i\|_2}\right)$ for all ${d \choose r}$ possible $\Lambda \in \{0, 1\}^d$ which are $r$-sparse and assign the maximizer to $\Lambda^*$.
\STATE \textbf{Return:} $Y^* = U \txt{ gnorm}\left(\txt{diag}\left(\Lambda^*\right)\Sigma V^{\top}\right)$
where $\txt{gnorm}(A)_i := A_i \cdot \frac{\sqrt{h}(\|A_i\|_2)}{\|A_i\|_2}$
for the $i^\txt{th}$ column of matrix $A$. 
\end{algorithmic}
\end{algorithm}

\begin{corollary}
Algorithm~\ref{alg:relu_svd} is fixed-parameter (in rank $r$) tractable polynomial time algorithm.
\end{corollary}
\begin{proof}
This statement follows directly from the efficiency of SVD and matrix multiplication operations, and the fact that we enumerate over $\Theta(d^r)$ possibilities.
\end{proof}


\subsection{Lower Bounding the Gap: Frobenius Approximation vs. $\NLRA$} \label{subsec:lower_bounding_SVD_gap}

We now use the developed theory to prove a lower bound on the sub-optimality of using the SVD to optimize the objective of Problem~\ref{prob:learning_prob_statement} when given $W$. We defer the proofs to the appendix.

Our main characterization theorem is as follows:

\begin{theorem}[Lower Bound on Suboptimality of SVD for $\NLRA$]\label{thm:subopt_lowerbound}
Recall the objective $\mathcal{R}(Y)$
from Problem~\ref{prob:learning_prob_statement},
where we require that $Y \in \mathbb{R}^{d \times m}$ is a rank-$r$ matrix.
Let $W \in \mathbb{R}^{d \times m}$. Define $\rho_{\sigma}^* \in \mathbb{R}^m$ as the correlations $\|\Sigma \txt{ diag}\left(\Lambda^*\right) V_i\|_2 / \|\Sigma V_i\|_2$, where $\Lambda^* \in \{0, 1\}^d$ is the solution to the optimization problem posed in Theorem~\ref{thm:relu_svd}. As shorthand, denote $Y(\rho)$ as the associated low-rank matrix for correlation vector $\rho \in \mathbb{R}^m$ (computed as described in Theorem~\ref{thm:relu_svd}). Denote $\rho_{\txt{SVD}}^*$ to be the optimal correlations in the case where we pick $\Lambda^*$ to correspond to the top-$r$ singular values, as in SVD.
Then, we have the following lower bound for the suboptimality of the SVD solution $Y_{\txt{SVD}}$:
$
     \frac{1}{d}(\mathcal{R}(Y_{\txt{SVD}}) - \mathcal{R}(Y(\rho_{\sigma}^*))) \geq \frac{1}{2d}\|w\odot \sqrt{h(\rho_{\txt{SVD}}^*)} - \rho_{\txt{SVD}}^*\|_2^2
$,
where $h$ is defined in Definition~\ref{def:hfunc}; $w = \begin{bmatrix} \|W_1\|_1 \cdots  \|W_m\|_2 \end{bmatrix}$ is a vector of column norms of $W$ with $W_i$ being the $i^\txt{th}$ column of $W$, and $\odot$ is the elmentwise product. We normalize by $d$ to get an average over the entry-wise error of the approximated output.
\end{theorem}

Corollary~\ref{lemma:svd_relusvd_relation} provides some intuition for Theorem~\ref{thm:subopt_lowerbound}:
\begin{corollary}[Relationship between SVD and ReLU SVD]\label{lemma:svd_relusvd_relation}
Suppose $W = U\Sigma V^{\top} \in \mathbb{R}^{d \times m}$.
Consider the solution for rank-$r$ ReLU SVD (given by $Y^*$) as described in Theorem~\ref{thm:relu_svd}. 
If $h(\rho)$ is replaced with $\rho^2$, and we always choose $\Lambda^*$ to correspond to the top $r$ singular values of $\Sigma$, then $Y^*$ is the standard SVD solution.
\end{corollary}

\begin{figure}
    \centering
    \includegraphics[scale=0.55]{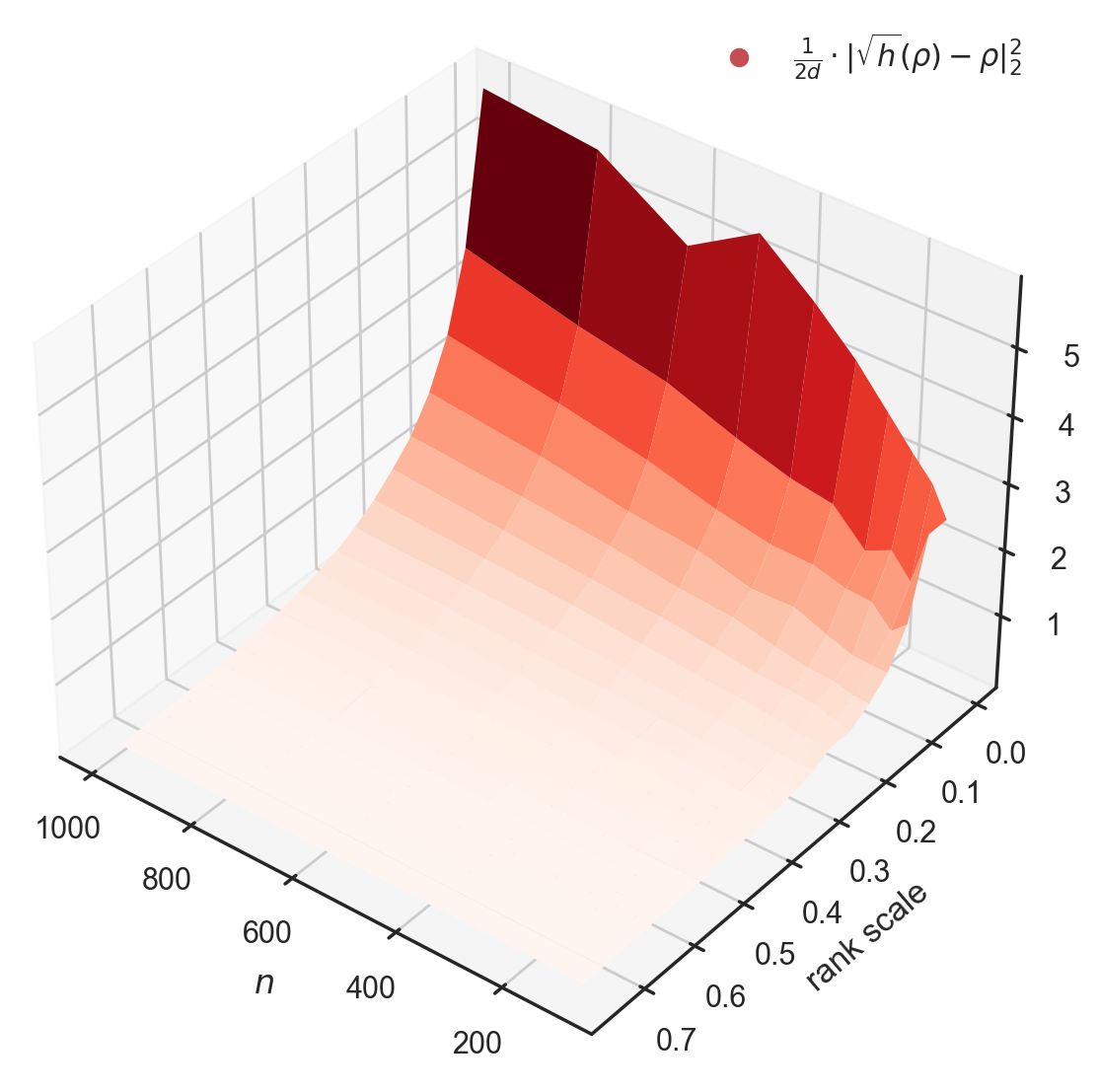}
    \caption{We plot the gap growth $\frac{1}{2d}\|\sqrt{h}(\rho) - \rho\|_2^2$ (see Theorem~\ref{thm:subopt_lowerbound}) for $W \in \R^{d \times m}$ with $m = n^{1.5}$ and $d = 0.2n$ with respect to the ReLU nonlinearity. Note the width is super-linear in dimension: $m = \Omega(n^{1 + \eps})$ for all $\eps > 0$. The entries of ground truth matrix $W$ are drawn from $\mathcal{N}(0, 1)$, and then we normalize $\|W_i\|_2 = 1, ~\forall i \in [m]$. We observe that the gap increases with increasing dimension and decreasing rank scale. Note that the behavior demonstrated matches the theoretical predictions: the gap increases as $\Theta((d^{1/2} \cdot (1 - \sqrt{r/d})^2)$ as per Corollary~\ref{cor:spherical_weights_dim_increase_rank_decrease}.}
    \label{fig:increasing_dim_decreasing_rank_gap}
\end{figure}

Using the lower bound in Theorem~\ref{thm:subopt_lowerbound}, we can now characterize the conditions on the full-rank matrix $W \in \R^{d \times m}$, where the solution to Problem~\ref{prob:learning_prob_statement} and the SVD solution are significantly different. First, smaller correlations $\rho_{\txt{SVD}}^*$ result in better solutions -- this case roughly corresponds to the columns of $W$ being approximately orthogonal (Corollary~\ref{cor:small_rho_larger_gap}, Remark~\ref{rem:orthogonality_intuition}).
When $\max_{i} \rho_{\txt{SVD}}^*(i) < 1$, we can prove that if the width $m$ is super-linear in $d$, the sub-optimality gap grows as $d$ increases, and furthermore the sub-optimality gap is monotone non-decreasing as $r$ decreases (Corollary~\ref{cor:dim_increase_rank_decrease}).
Finally, we prove a stronger version of Corollary~\ref{cor:dim_increase_rank_decrease} under the assumption of Gaussian distributed columns of $W \in \R^{d \times m}$, and recover the actual dependence on the rank scale $r/d$:

\begin{corollary}[Spherical Weights]\label{cor:spherical_weights_dim_increase_rank_decrease}
Suppose the columns of $W \in \R^{d \times m}$ are drawn from the uniform distribution over the surface of the unit sphere\footnote{This assumption is reasonable, given the many similar init distributions used in practice (e.g., He and Glorot inits \citep{he2015delving, glorot2010understanding}).}. Consider target rank $r \leq d$. Assume that $r \gg \Theta(\log(m))$. Then with probability at least constant,
$
\rho_{\max} \leq \sqrt{\frac{r}{d}}
$,
and thus
$
    \frac{1}{d}\left(\mathcal{R}(Y_{\txt{SVD}})) - \mathcal{R}(Y(\rho_{\sigma}^*))\right) \geq \Omega\left(d^{\delta}\cdot \left(\frac{1}{\pi} - \frac{1}{2} \sqrt{\frac{r}{d}}\right)^2\right)
$,
assuming that $m = d^{1 + \delta}$ for some $\delta > 0$
and that $\sqrt{\frac{r}{d}} < \frac{2}{\pi}$.
\end{corollary}
Therefore, in the case of spherical $W$, the sub-optimality gap increases as either the dimension increases (when width is super-linear in dimension) or as the rank scale decreases. 

\begin{proof}[Sketch]
\looseness=-1
We make use of Theorem~\ref{thm:subopt_lowerbound} to lower bound the gap -- we achieve this by bounding each term in a sum over the $m$ columns uniformly. This step reduces to upper bounding the size of $\rho_{\max}$. Intuitively, the dependence on $\sqrt{r/d}$ comes from the fact that the correlations $\rho_i$ are $m$ random variables that are near-Gaussian in behavior (they can be expressed as $\ell_2$ norms of a sum of $r$ i.i.d. Gaussian entries, normalized by the $\ell_2$ norm of a $d$-dimensional Gaussian). Thus, taking the expectation of the max over these variables produces a value $\sim \Theta(\sqrt{\log(m)/d})$. By making use of a concentration argument analogous to the standard Johnson-Lindenstrauss Lemma \citep{vershynin2018high}, if $r > \Omega(\log m)$, we are able to obtain the desired result. The super-linear requirement on $m$ comes from the fact that we are normalizing a sum over $m$ (the layer-width) columns by input dimension $d$.
\end{proof}

\section{Experiments and Discussion}\label{sec:experiments}


We present some empirical results for our low-rank initialization scheme for training EfficientNets \citep{tan2019efficientnet} on the ImageNet dataset \citep{ILSVRC15} with stochastic gradient descent and momentum, with tuned parameters and learning rate schedule. We study low-rank variants of these networks for various choices of \textit{rank scale} -- for parameter matrix $W \in \R^{d \times m}$, the rank scale is the fraction of $\min(d, m)$ that we require for our low-rank factorization $UV^{\top}$. We view the choice of rank scale as a trade-off parameter between computation and accuracy since in our settings, lower rank uniformly means worse generalization (though this may be false for other tasks and datasets).
We compare the following initialization methods: 1) Baseline Low-Rank -- apply the full-rank init distribution for $W$ to low-rank factors $U \in \R^{m \times r}, V \in \R^{d \times r}$; 2) Spectral (Definition~\ref{def:spectral_init}); 3) LFAI-Adam (Algorithm~\ref{alg:lfai_gradient}), implemented with Adam \citep{kingma2014adam}; 4) LFAI-WS-Adam (Algorithm~\ref{alg:lfai_ws_gradient}). 


\subsection{Details on Main Experimental Setup}
We train ResNet50 \citep{he2016deepresidual} and EfficientNet \citep{tan2019efficientnet} models with stochastic gradient descent and momentum on the $2012$ ImageNet dataset \citep{ILSVRC15}, with tuned parameters and learning rate schedule. The low-rank version of this model simply replaces the convolution layers with low-rank convolutions, as described in \citep{khodak2021initialization}. The full-rank weight initialization is a truncated normal distribution $\mathcal{N}(0, 1/d)$, where $d$ is the number of input units for the layer, and where ``truncation'' refers to discarding and re-sampling any samples which are more than two standard deviations from the mean. 

We also consider two kinds of regularization on the objective:
\begin{enumerate}
    \item Weight decay: This method is the standard Frobenius norm regularization on the weights of the layers. In the low-rank setting, instead of penalizing $\|W\|_F^2$, we penalize $\|U\|_F^2 + \|V\|_F^2$.
    \item Frobenius decay: This regularization approach is demonstrated by \citet{khodak2021initialization} to outperform weight decay in several settings they consider. Instead of separately regularizing the low rank factors, this approach penalizes $\|UV^T\|_F^2$.
\end{enumerate}
We tune the regularization strength separately for both approaches and report the performance of the best regularization strength. Tuning the Frobenius decay regularization strength did not succeed for the EfficientNet models due to divergence during training. Thus we only report the weight decay results for the EfficientNet models. For ResNet-$50$, for Frobenius decay, we use a regularization strength of $0.3$. For ResNet-$50$ weight decay, we use a regularization strength of $10^{-4}$. For the EfficientNet models with weight decay, we use a regularization strength of $10^{-5}$.
We compare across multiple choices of rank scale $\frac{r}{d}$: $0.05, 0.1, 0.15, 0.2$. 

We run each experiment three times for three different random seeds and average the results and include the standard error up to one standard deviation on the mean performance. The differences across each experiment instance are due to the changes in random seeds used in both sampling at initialization and for sampling batches during optimization. We trained the models on TPU hardware.

For $\NLRA$-based approaches, we choose the standard Gaussian as the input distribution. However, other choices are also feasible:
\begin{remark}[On the Choice of Input Distribution]\label{rem:choice_of_input}
We minimize the function approximation loss over the Gaussian distribution. 
In practice, we may view this choice as a hyper-parameter to tune. For instance, another reasonable (but computationally expensive) approach is to minimize the function approximation loss over the real data distribution (after being transformed by the previous input layers). In our experiments, we did not notice much difference by switching to this initialization distribution, but it is plausible that in other problem settings other input distributions might perform better.
\end{remark}

\subsubsection{Model Architecture}

The EfficientNet architecture for b$0$, b$3$, and b$7$ is standard and available in libraries such as Tensorflow \citep{abadi2015tensorflow}. See for instance \href{https://github.com/keras-team/keras/blob/v2.7.0/keras/applications/efficientnet.py}{https://github.com/keras-team/keras/blob/v2.7.0/keras/applications/efficientnet.py}. For EfficientNet-b$9$, we define the width scale parameter to be $3.0$, and the depth-scale parameter to be $3.2$.
For both architectures, we used the Swish nonlinearity \citep{ramachandran2017searching}.
For ResNet-$50$, we also simply use the standard available architecture in Tensorflow.

\subsubsection{Hyperparameters for Layerwise Initialization}

In our experiments, we implement both LFAI-Adam and LFAI-WS-Adam using the Adam optimizer \citep{kingma2014adam}. Since our networks consist of low-rank convolutional layers (see \citet{khodak2021initialization} for guidance on how to efficiently define a low-rank convolutional layer), sample $1000$ Gaussian vectors in the shape $(64, 64, \txt{num. input filters})$ -- we arbitrarily choose $64 \times 64$ patches since the specific dimension of the ``image'' does not matter too much, and we want to avoid making the dimension of our training problem too large. The number of input filters is determined by the layer and the architecture -- for the first layer, there are $3$ input filters. 

For Adam, we set the learning rate to $5 \times 10^{-3}$, the batch size to $512$, and the number of steps per epoch to $128$ after hyper-parameter searching over these parameters to determine the quickest convergence rate. With these parameters, Adam converges to the optimum for all low-rank layerwise optimization problems we tried within $6$ epochs.

We reported results both for initializing Adam with spectral initialization (LFAI-WS-Adam) and with baseline initialization (LFAI-Adam) -- almost universally LFAI-WS-Adam is better. 

\subsubsection{Hyperparameters for Post-Initialization Optimization}

For ResNet-$50$ and each EfficientNet model, we trained the network for $62000$ epochs using Stochastic Gradient Descent (SGD) with Momentum. We set the batch size to $4096$. We set the momentum parameter to $0.9$ and the learning rate schedule to follow a linear warm-up schedule for $1560$ steps (choose values for the learning rate from $0$ to an initial rate of $1.6$), followed by a cosine curve with an initial rate of $1.6$, decaying over the remaining $60440$ steps (see the CosineDecay learning rate schedule in Tensorflow). We did not modify this training scheme across our experiments for simplicity (our goal was to compare the performance of different init schemes rather than to attain the optimal performance), explaining why our full-rank EfficientNet numbers do not match the numbers in the original EfficientNet paper \citep{tan2019efficientnet}.

\subsection{Results}
It turns out that either LFAI-WS-Adam or LFAI-Adam typically outperform the other initialization schemes across multiple rank scales and architecture widths. On average across all experiments (including on ResNet \citep{he2016deepresidual}, see Section~\ref{subsec:more_experiment_results}), our method gains on the order of $0.3\%$ in accuracy, though the gain is larger for smaller rank scales and larger width networks. These
results are also significant -- we demonstrate a clear separation of our approach compared to others in $1$-standard deviation confidence intervals, and our method tends to have lower variance in performance. 
Furthermore, we also find empirical evidence of our theoretical claims in Section~\ref{sec:properties_nonlinear_lowrank}, despite the fact that we use the Swish activation while our theory was ReLU-based: we see the trend that as the rank-scale decreases, or as the widths of the networks increase (higher number EfficientNets correspond to larger widths), we see improvement in the top-$1$ accuracy gain, see Table~\ref{table:effnet_results}.

While Frobenius Decay outperforms Weight Decay for smaller rank scales, the advantage erodes for larger rank scales. This effect was not observed by \citep{khodak2021initialization},
possibly because we study larger models and datasets. We note that our method improves over other initialization methods regardless of the initialization scheme.

We also note that in our EfficientNet experiments (Table~\ref{table:effnet_results}), the effect of width does not appear to be as strong as the effect of the rank scale -- loosely using the intuition built by our theory, this empirical result makes sense: We are likely not in a setting where the width is super-linear in the input dimension. In fact, we are likely in the linear regime, where $m = \mc{O}(d)$. Since the networks are deep, it is difficult to directly use the $1$-hidden-layer metaphor from our theory -- nevertheless, for many of the layers in the network, the difference between input and output dimension is typically not more than a factor of $2$ -- even for the EfficientNet architecture which we scale increasingly with depth. Thus, we do not expect as dramatic gains when we scale up the widths of the networks, though we do see some mild gain. On the other hand, the effect of the smaller rank scale is quite apparent in our experiments.

It would be interesting to develop a more thorough characterization of the impact of the ratios of the inputs and outputs of layers in a deep architecture on the effectiveness of the LFAI-Adam framework -- it seems plausible that we would have to take into account some notion of average ratio between input and output sizes, and possibly the structure of the architecture itself as well.

\begin{figure}
    \centering
    \includegraphics[scale=0.63]{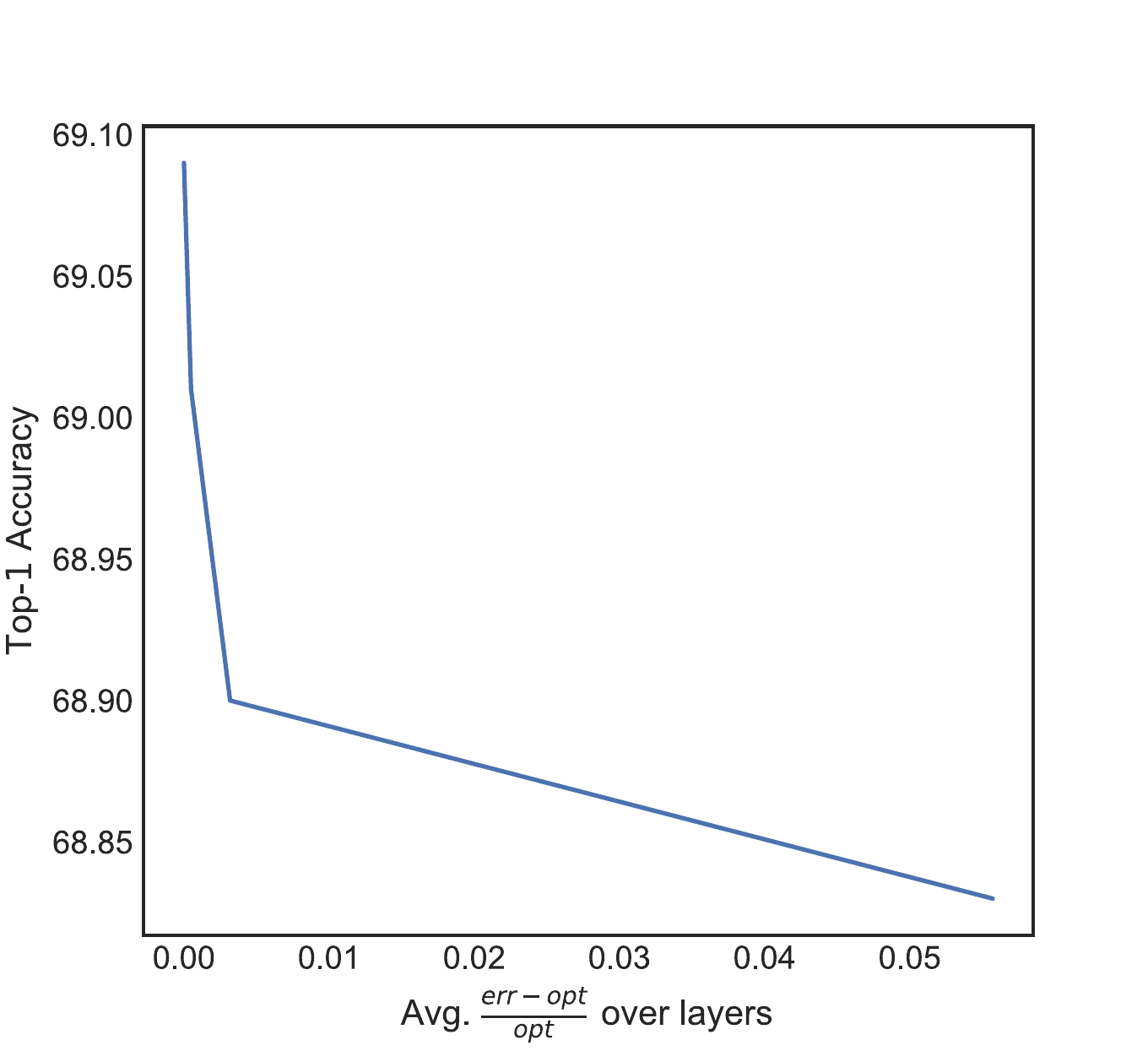}
    \caption{We vary the number of optimization steps used to implement WS-\algnameshort and plot the average normalized error across layers against the average downstream validation top-$1$ accuracy for training an EfficientNet-b$7$ model on ImageNet. Decreasing the function approximation error at init is beneficial for top-$1$ accuracy.}
    \label{fig:downstream_generalization_from_approx_at_init}
\end{figure}

In Table~\ref{table:resnet_results}, we present our results for the ResNet-50 architecture \citep{he2016deepresidual}. In Tables~\ref{table:b9_compare}, \ref{table:b7_compare}, \ref{table:b3_compare},  \ref{table:b0_compare}, \ref{table:resnet50_weightdecay_compare}, and \ref{table:resnet50_frobdecay_compare}, we summarize the performance of our best initialization method compared to the others.

\begin{table}
\centering
\caption{Comparison of LFAI-WS-Adam Across Rank Scales for EfficientNet-b9. We show by how many percent points the best performing method, LFAI-WS-Adam, beats the other methods, for the Weight Decay regularization.}
\label{table:b9_compare}
\vskip 0.15in
\begin{tabular}{|l|l|l|}
\hline
 Rank Scale & Top-$1$ Accuracy for LFAI-WS-Adam  \\  \hline
 $0.05$ & $ 66.90 \pm 0.11 $  (best by $0.36$) \\ \hline
 $0.1$ & $ 72.28 \pm  0.04$ (best by $0.21$) \\ \hline
 $0.15$ & $ 73.31 \pm 0.04$ (sub-optimal) \\ \hline
 $0.2$ & $ 74.01 \pm 0.11 $ (overlaps w/best) \\ \hline
 Full-Rank & $79.62 \pm 0.05$  \\ \hline
\end{tabular}
\end{table}

\begin{table}
\centering
\caption{Comparison of LFAI-WS-Adam Across Rank Scales for EfficientNet-b7. We show by how many percent points the best performing method, LFAI-WS-Adam, beats the other methods, for the Weight Decay regularization.}
\label{table:b7_compare}
\vskip 0.15in
\begin{tabular}{|l|l|l|}
\hline
 Rank Scale & Top-$1$ Accuracy for LFAI-WS-Adam  \\  \hline
 $0.05$ & $59.63 \pm 0.37$  (best by $0.63$) \\ \hline
 $0.1$ & $69.09 \pm 0.16$ (best by $0.20$) \\ \hline
 $0.15$ & $71.74 \pm 0.02$ (best by $0.07$) \\ \hline
 $0.2$ & $72.72 \pm 0.03$ (best by $0.27$) \\ \hline
 Full-Rank & $78.70 \pm 0.02$  \\ \hline
\end{tabular}
\end{table}

\begin{table}
\centering
\caption{Comparison of LFAI-Adam Across Rank Scales for EfficientNet-b3. We show by how many percent points the best performing method, LFAI-Adam, beats the other methods, for the Weight Decay regularization.}
\label{table:b3_compare}
\vskip 0.15in
\begin{tabular}{|l|l|l|}
\hline
 Rank Scale & Top-$1$ Accuracy for LFAI-Adam  \\  \hline
 $0.05$ & $38.22 \pm 0.28$  (overlaps w/best) \\ \hline
 $0.1$ & $56.96 \pm 0.19$ (overlaps w/best) \\ \hline
 $0.15$ & $63.55 \pm 0.16$ (best by $0.53$) \\ \hline
 $0.2$ & $66.32 \pm 0.02$ (best by $0.27$) \\ \hline
 Full-Rank & $76.63 \pm 0.19$  \\ \hline
\end{tabular}
\end{table}

\begin{table}
\centering
\caption{Comparison of LFAI-Adam and LFAI-WS-Adam Across Rank Scales for EfficientNet-b0. We show by how many percent points LFAI-WS-Adam compares against the other methods, for the Weight Decay regularization.}
\label{table:b0_compare}
\vskip 0.15in
\begin{tabular}{|l|l|l|}
\hline
 Rank Scale & Top-$1$ Accuracy for LFAI-Adam or LFAI-WS-Adam  \\  \hline
 $0.05$ & $32.39 \pm 0.26$  (best by $1.17$) \\ \hline
 $0.1$ & $47.84 \pm 0.24$ (suboptimal) \\ \hline
 $0.15$ & $55.19 \pm 0.10$ (best by $0.20$) \\ \hline
 $0.2$ & $59.38 \pm 0.18$ (suboptimal) \\ \hline
 Full-Rank & $74.50 \pm 0.13$ \\ \hline
\end{tabular}
\end{table}

\begin{table*}[t] 
\centering
\caption{Average Top-$1$ accuracy on ResNet-50 with different regularization methods. Note that our method tends to outperform the baselines regardless of whether Weight Decay or Frobenius Decay is used. For each rank scale, we display in bold all methods whose confidence intervals overlap.}
\label{table:resnet_results}
\vskip 0.15in
\begin{tabular}{|c|c|c|c|}
\hline
 Rank Scale & Method & Weight Decay  & Frobenius Decay \\ \hline
 \multirow{4}{*}{0.05} & Baseline &  $56.92 \pm 0.14$ & $57.73 \pm 0.23$ \\ \cline{2-4}
 & Spectral &  $56.28 \pm 0.39$ & $57.42 \pm 0.38$ \\ \cline{2-4}
 & LFAI-Adam & $56.55 \pm 0.04$ & $57.61\pm 0.15$ \\ \cline{2-4}
 & \textbf{LFAI-WS-Adam} & $\mathbf{57.37 \pm 0.08}$ & $\mathbf{58.22 \pm 0.31}$ \\ \hline
 \multirow{4}{*}{0.10} & Baseline &  $64.09 \pm 0.12$ & $65.61 \pm 0.09$\\ \cline{2-4}
 & Spectral &  $64.06 \pm 0.19$ & $65.69 \pm 0.12$\\ \cline{2-4}
 & LFAI-Adam & $64.02 \pm 0.17$& $65.55 \pm 0.05$  \\ \cline{2-4}
 & \textbf{LFAI-WS-Adam} & $\mathbf{64.74 \pm 0.04}$ & $\mathbf{65.93 \pm 0.05}$ \\ \hline
 \multirow{4}{*}{0.15} & Baseline &  $67.58 \pm 0.15$ & $\mathbf{68.15 \pm 0.20}$\\ \cline{2-4}
 & Spectral &  $66.95 \pm 0.16$ &$\mathbf{68.21 \pm 0.08}$ \\ \cline{2-4}
 & LFAI-Adam & $67.24 \pm 0.26$& $67.13 \pm 0.51$ \\ \cline{2-4}
 & \textbf{LFAI-WS-Adam} & $\mathbf{67.74 \pm 0.03}$ & $\mathbf{67.83 \pm 0.42}$ \\ \hline
  \multirow{4}{*}{0.20} & \textbf{Baseline} &  $\mathbf{69.04 \pm 0.33}$ & $69.22 \pm 0.05$ \\ \cline{2-4}
 & Spectral &  $68.59 \pm 0.19$ & $69.04 \pm 0.14$ \\ \cline{2-4}
 & LFAI-Adam & $68.93 \pm 0.20$ & $68.87 \pm 0.10$  \\ \cline{2-4}
 & \textbf{LFAI-WS-Adam} & $\mathbf{69.37 \pm 0.07}$& $\mathbf{69.47 \pm 0.09}$  \\ \hline
 1.0 & Full-Rank & $78.1$ & $78.1$\\ \hline
\end{tabular}
\end{table*}

\begin{table}
\centering
\caption{Comparison of LFAI-WS-Adam Across Rank Scales for ResNet-50. We show by how many percent points the best performing method, LFAI-WS-Adam, beats the other methods, for the Weight Decay regularization.}
\label{table:resnet50_weightdecay_compare}
\vskip 0.15in
\begin{tabular}{|l|l|l|}
\hline
 Rank Scale & Top-$1$ Accuracy for LFAI-WS-Adam  \\  \hline
 $0.05$ & $57.37 \pm 0.08$  (best by $0.45$) \\ \hline
 $0.1$ & $64.74 \pm 0.04$ (best by $0.65$) \\ \hline
 $0.15$ & $67.74 \pm 0.03$ (best by $0.16$) \\ \hline
 $0.2$ & $69.37 \pm 0.07$ (best by $0.33$) \\ \hline
 $0.25$ & $70.37 \pm 0.06$ (best by $0.25$) \\ \hline
 Full-Rank & $78.1$  \\ \hline
\end{tabular}
\end{table}

\begin{table}
\centering
\caption{Comparison of LFAI-WS-Adam Across Rank Scales for ResNet-50. We show by how many percent points the best performing method, LFAI-WS-Adam, beats the other methods, for the Frobenius Decay regularization.}
\label{table:resnet50_frobdecay_compare}
\vskip 0.15in
\begin{tabular}{|l|l|l|}
\hline
 Rank Scale & Top-$1$ Accuracy for LFAI-WS-Adam \\  \hline
 $0.05$ & $58.22 \pm 0.31$ (best by $0.49$) \\ \hline
 $0.1$ & $65.93 \pm 0.05$ (best by $0.24$) \\ \hline
 $0.15$ & $67.83 \pm 0.42$ (sub-optimal) \\ \hline
 $0.2$ & $69.47 \pm 0.09$ (best by $0.25$) \\ \hline
 $0.25$ & $70.16 \pm 0.09$ (best by $0.19$) \\ \hline
 Full-Rank & $78.1$  \\ \hline
\end{tabular}
\end{table}

\subsection{Downstream Generalization Error}
While existing literature has studied the effects of various initializations on optimization, no results as far as we are aware connect the choice of initialization to downstream generalization performance, though this effect has been demonstrated empirically. 

Thus, we empirically justify the benefits of function approximation at initialization for low-rank networks. 
We perform the following experiment: We fix rank scale $0.10$, EfficientNet-b$7$ and initialization algorithm LFAI-WS-Adam, and investigate the effect of layerwise function approximation at initialization on downstream generalization. We implement layerwise function approximation with Adam, as mentioned above.  We train the network on ImageNet training data starting from the initialization produced after $0-6$ epochs of running Adam starting from the spectral initialization, and measure the top-$1$ validation accuracy. For each number of epochs of training, we run the experiment $3$ times and take the average.

Then, on the $x$-axis, for each number of initialization training epochs, we plot the average multiplicative error \textit{across} the $107$ layers of EfficientNet-b$7$, paired with the corresponding downstream generalization error. In particular, if the optimum value of our objective (defined in Problem~\ref{prob:learning_prob_statement}) at a given layer (at init) is $\txt{opt}$, and we have reached $(1 + \eps_{\txt{layer}~i})\cdot \txt{opt}$, then we record the average $\eps$ across all layers for that number of epochs, and plot alongside it the corresponding average over downstream top-$1$ validation accuracy. We find that after $4$ epochs, training converges and the optimization procedure has reached the optimum. 

We plot our results in Figure~\ref{fig:downstream_generalization_from_approx_at_init}, and observe that as LFAI-WS-Adam optimizes, accuracy improves, demonstrating that optimal function approximation at initialization improves downstream generalization.

\section{Conclusion}\label{sec:conclusion}
We introduced a novel low-rank initialization framework for deep learning: Algorithms~\ref{alg:lfai_gradient} and ~\ref{alg:lfai_ws_gradient} empirically tend to outperform the existing method of choice, spectral initialization, while being essentially as efficient to implement. We view our approach as a simple drop-in substitute to spectral initialization that practitioners can try to increase performance. While we do not expect huge performance increases, since the method is simple and does not require significantly more compute or memory, Table~\ref{table:effnet_results} demonstrates a relatively reliable performance increase in generalization accuracy.

This paper also contributes to the understanding of the mechanism behind successful low-rank initialization methods for deep learning: a critical component of methods like spectral initialization and our $\NLRA$ framework is the quality of the \textit{function approximation} of the full-rank initialized parameters, rather than parameter approximation. We demonstrate this with empirical results showcasing the positive correlation between decreasing nonlinear low-rank approximation error and decreasing generalization error, and with theoretical guarantees and intuition for when our methods should outperform classic approaches (high input dimension together with larger network width and  smaller target ranks) (Theorem~\ref{thm:subopt_lowerbound} and Corollary~\ref{cor:spherical_weights_dim_increase_rank_decrease}). We also resolve an open theoretical question by providing a provably polynomial time fixed-parameter tractable algorithm for solving ReLU low-rank function approximation in the rank $r$ (Algorithm~\ref{alg:relu_svd} and Theorem~\ref{thm:relu_svd}).  

We close with a few directions for future work: It would be interesting to 1) provably characterize the impact of initialization schemes on downstream generalization, low-rank or otherwise; 2) identify optimal \textit{distributions} to sample low-rank weights from without optimization; 3) discover better low-rank training methods; 4) extend our theory beyond $1$-hidden-layer ReLU networks to other activations, depths, and architectures; 5) extend our initialization framework to ``efficiently-parameterized'' networks beyond low-rank; 6) determine the complexity of the nonlinear low-rank approximation problem when the rank is not constant, as well as understand the sample complexity and computational complexity of the problem when one is only provided sample access to the weights.

\FloatBarrier

\bibliography{references}
\bibliographystyle{icml2022}

\newpage

\appendix
\section{Proofs}
\label{appendix:proofs}

\subsection{Background on Hermite Analysis}\label{subsec:hermite}

We present some theory about the Hermite polynomial basis which is useful in the analysis. We take the definition and required basic facts from \citet{ODonnell14}. 

\begin{definition}[Gaussian Space]\label{def:gaussian_space}
$L_2(\mc{N}(0, 1))$ is Hilbert space with respect to the Hermite orthogonal polynomial basis.
\end{definition}

\begin{definition}[Hermite orthogonal polynomial basis]
The Hermite basis is an orthogonal basis over $L_2(\mc{N}(0, 1))$. In particular, we can write 
\begin{align*}
f(a) &= \sum_{\ell = 0}^{\infty} c_{\ell} H_{\ell}(a)
\end{align*}
where $H_{\ell}(a)$ is the ${\ell}^{th}$ Hermite basis function, and $c_{\ell} = \E{a}{f(a) H_{\ell}(a)}$. 
We define $$ H_0(a) = 1, \quad H_1(a) = a , $$ and compute the rest by applying Gram-Schmidt over the function space. We have the following definition for the $\ell^{th}$ Hermite basis function:
\begin{align*}
    H_{\ell}(a) := \frac{1}{\sqrt{{\ell}!}}\frac{(-1)^{\ell}}{\varphi(a)} \frac{d^{\ell}}{da^{\ell}}\varphi(a) .
\end{align*}
We also have the recurrence relation
\begin{align*}
    H_{{\ell} + 1}(a) &= \frac{1}{\sqrt{{\ell} + 1}}\left(aH_{\ell}(a) - \frac{d}{da}H_{\ell}(a)\right)
\end{align*}
and derivative formula
\begin{align*}
    \frac{d}{da}H_{\ell}(a) &= \sqrt{{\ell}}H_{{\ell} - 1}(a) .
\end{align*}
\end{definition}

The following important lemma provides a rule for calculating $\mathbb{E}_{a, a'}\left[f(a)f(a')\right]$ when $a, a'$ are correlated standard Gaussian random variables. 
\begin{lemma} \label{hermite_trick} 
Let $a, a'$ be standard Gaussian random variables with correlation $\rho$. Then, we have
\[
  \E{a, a'}{H_{\ell}(a)H_{\ell'}(a')} = \begin{cases} \rho^{\ell} & \txt{ if } {\ell} = {\ell'} \\ 0 & \txt{ otherwise.} \end{cases}
\]
and
\begin{align*}
    \E{a, a'}{f(a)f(a')} &= \sum_{{\ell}, {\ell'} = 1}^{\infty} c_{\ell}c_{\ell'} \E{a, a'}{H_{\ell}(a)H_{\ell'}(a')}
    \\
    &= \sum_{{\ell} = 0}^{\infty} c_{\ell}^2 \rho^{\ell} .
\end{align*}
\end{lemma}

\subsection{Hermite Decomposition of ReLU}

\citet{goel2019time} derives properties of the Hermite expansion for the univariate ReLU:
\begin{lemma}[Hermite Expansion Properties for Univariate ReLU \citep{goel2019time}]
Let $\left\{c_i\right\}_{i = 0}^{\infty}$ be the Hermite coefficients for ReLU. Then, 
\[
    c_k = \begin{cases} 1/\sqrt{2\pi} &\txt{ if } k = 0 \\
                        \frac{1}{2} &\txt{ if } k = 1 \\
                        \frac{1}{\sqrt{2\pi k!}}\left(H_k(0) + kH_{k - 2}(0)\right) &\txt{ if } k \geq 2
          \end{cases}
\]
\end{lemma}

Using the above properties, we can derive the explicit form of the Hermite coefficients for the ReLU activation:

\begin{lemma}[Hermite Coefficients for ReLU]\label{hermite_coeff_relu}
Let $\left\{c_i\right\}_{i = 0}^{\infty}$ be the Hermite coefficients for ReLU. Then, 
\[
    c_k = \begin{cases} 1/\sqrt{2\pi} &\txt{ if } k = 0 \\
                        \frac{1}{2} &\txt{ if } k = 1 \\
                        0 &\txt{ if } k = 2m + 1, m \geq 1 \\
                        \sqrt{\frac{1}{2\pi} \frac{1}{4^m} \cdot {2m \choose m} \cdot \frac{1}{(2m - 1)^2}} &\txt{ if } k = 2m, m \geq 1
          \end{cases}
\]
\end{lemma}
\begin{proof}
First we show that $c_k = 0$ for odd $k > 1$. This is easy to check since if $k$ is odd, so is $k - 2$ and checking that there is no constant term for odd Hermite polynomials $H_k$ yields that the whole expression is $0$.

For the rest of the even terms, plug in the standard formula 
\[
H_{2m}(0) = \left(-1\right)^m \frac{(2m)!}{m! \cdot 2^m}
\]
to the recurrence given in the expansion properties and simplify using ${2m \choose m} = \frac{(2m)!}{m! m!}$.
\end{proof}

\begin{lemma}[Hermite Analysis of ReLU Correlation]\label{lem:hermite_analysis_correlation}
Suppose we have univariate standard Gaussians $g_1, g_2$ which are $\rho$-correlated ($\rho \in [0, 1]$). Let $\sigma(x) = \max(0, x)$ be the ReLU activation. Then,
\begin{align}
\begin{split}
  &2\E{g_1, g_2}{\sigma(g_1)\sigma(g_2)} 
  \\
  &= \frac{1}{\pi}\left(1 + \frac{\pi}{2}\rho + \sum_{\ell = 1}^{\infty} \frac{1}{4^{\ell}} \cdot {2\ell \choose \ell} \cdot \frac{1}{(2\ell - 1)^2} \rho^{2\ell}\right)  
  \\
  &=: \sqrt{h}(\rho) \label{def:hfunc}
\end{split}
\end{align}
\end{lemma}
\begin{proof}
Apply Lemma~\ref{hermite_trick} and Lemma~\ref{hermite_coeff_relu} and simplify the algebra. 
\end{proof}

\begin{figure}
    \centering
    \includegraphics[scale=0.58]{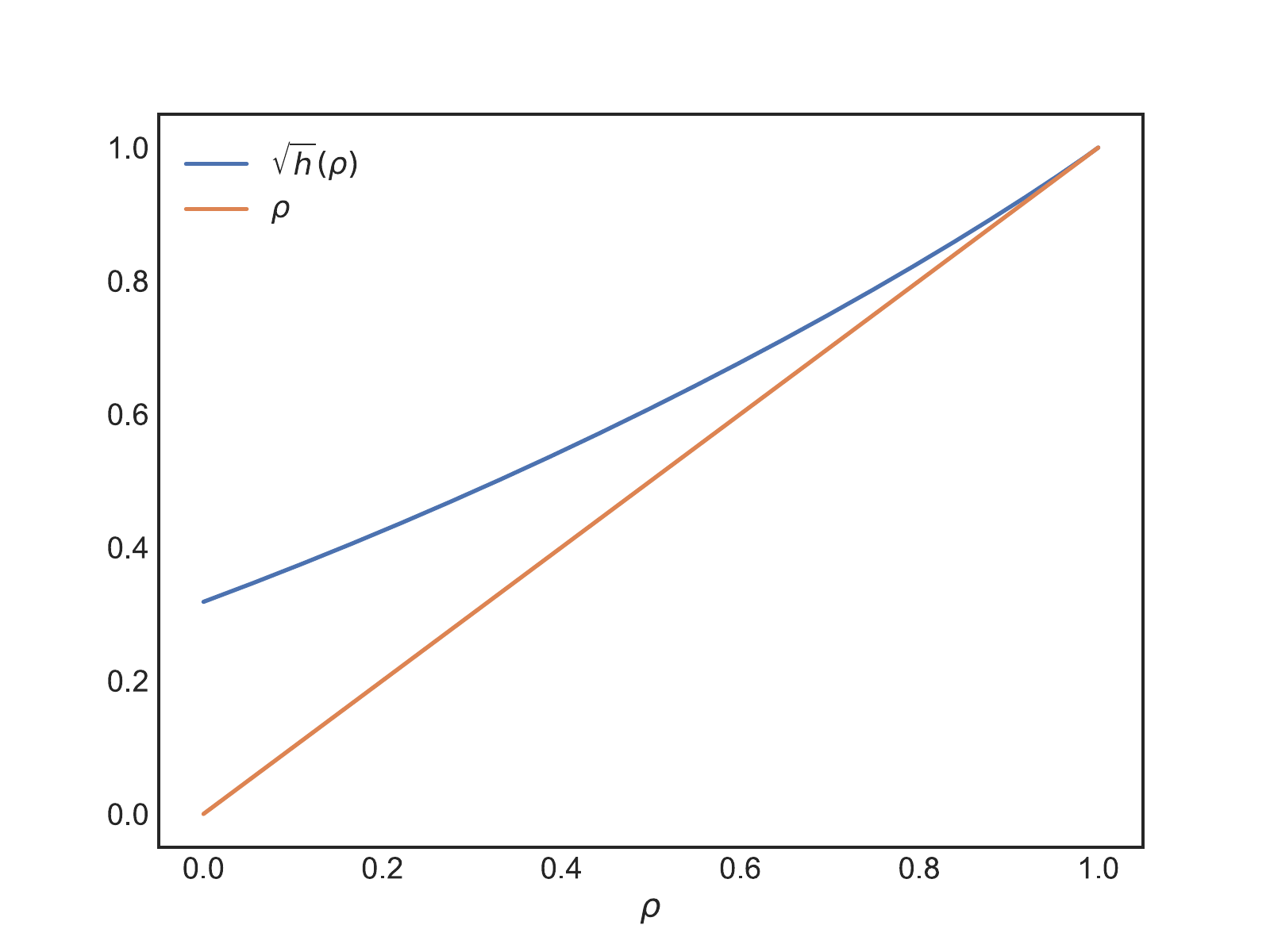}
    \caption{$\sqrt{h(\rho)}$ (see Definition~\ref{def:arccos_kernel}) plotted against the linear function $\rho$. Here, $\rho$ is a correlation between the inputs to the arc-cosine kernel.}
    \label{fig:arcccos_vs_linear}
\end{figure}

\begin{lemma}[Convexity and Monotonicity of $\sqrt{h}$] \label{lem:h_convex_monotone}
The function $\sqrt{h}$ defined by Definition~\ref{def:hfunc}
is both convex and monotone increasing on $[0, 1]$, and is also bounded between $[0, 1]$ for inputs in $[0, 1]$.
\end{lemma}
\begin{proof}
First we compute the derivative:
\[
\frac{d\sqrt{h}(\rho)}{d\rho} = \frac{1}{2} + \sum_{\ell = 1}^{\infty} \frac{1}{4^{\ell}}{2\ell \choose \ell}\frac{2\ell}{(2\ell - 1)^2} \rho^{2\ell -1}
\]
Observe the derivative is positive for $\rho \in [0, 1]$; thus $\sqrt{h}(\rho)$ is monotone increasing. Since $\sqrt{h}(\rho)$ is a positive combination of convex functions (linear functions and even powers are convex), it is also convex. To prove it is bounded, using monotonicity, we only need consider the extremes at $\rho = 0, 1$. We have $h(0) = \frac{1}{\pi}$, and we have (using the closed form given by Definition~\ref{def:arccos_kernel})
\[
h(1) = \frac{\left(\sqrt{1 - 1^2} + \left(\pi - 0\right)\cdot 1\right)}{\pi} = 1
\]
which proves the statement.
\end{proof}

\begin{remark}[Convexity and Monotonicity of C-Maps]\label{rem:cmap_convexity}
It is worth noting that the analysis in \citet{martens2021rapid} proves that a wide variety of related functions to $\sqrt{h}$ are convex and monotone.
\end{remark}

\subsection{Problem Statement Remarks}

We also have the following two remarks regarding our framing of the problem.
\begin{remark}[Recovering Low-Rank Factors]\label{rem:factorizing_Y}
Note that we have framed the learning problem in terms of recovering a low-rank matrix $Y^*$. To fulfill the promise of improved computational efficiency, the original motivation for considering this learning problem, we would apply a final post-processing step after learning $Y^*$ by exactly decomposing it into the product of two rank $r$ matrices using SVD.
\end{remark}

\begin{remark}[Convolutional Architecture]\label{rem:conv_arch}
In this problem we only consider fully-connected architecture. However, since a convolution is a linear operator, we can make use of results which efficiently learn one-layer convolutional ReLU networks \citep{du2018gradient} to first recover the convolutional filters before reducing to our low-rank recovery procedure.
\end{remark}




\subsection{Proofs for Section~\ref{sec:properties_nonlinear_lowrank}}\label{subsec:proofs_sec3}

We recall the following useful definition:

\begin{definition}[ReLU Kernel: $1^{st}$-Order Arc-Cosine Kernel]
We define a function known in the literature as the first-order arc-cosine kernel \citep{cho2009kernel}, and is defined by 
\[
k(x, y) := \|x\|_2 \|y\|_2 \cdot \sqrt{h}(\rho_{xy})
\]
where
\[
\rho_{xy} := \frac{x^{\top}y}{\|x\|_2\|y\|_2}
\]
and
\[
\sqrt{h}(\rho_{xy}) = \frac{\left(\sqrt{1 - \rho_{xy}^2} + \left(\pi - \cos^{-1}\left(\rho_{xy}\right)\right)\rho_{xy}\right)}{\pi}
\]
This function will be very relevant to our analysis of ReLU SVD.
\end{definition}

We provide an accompanying Hermite expansion of $\sqrt{h}$ in Lemma~\ref{lem:hermite_analysis_correlation}.

\subsubsection{Characterizing the ReLU SVD}

In the following theorem, we take advantage of specific properties of the ReLU activation and its associated kernel to come to another understanding of the objective in Problem~\ref{prob:learning_prob_statement}.
\begin{theorem}[ReLU SVD]
Consider the goal of finding the optimal rank-$r$ solution to the objective $\mc{R}(Y)$ in Problem~\ref{prob:learning_prob_statement} with known $W \in \R^{d \times m}$. 
Suppose we solve the following problem:
\[ 
\Lambda^* = \argmax_{\substack{\Lambda \in \{0, 1\}^{d};~\|\Lambda\|_1 = r}} \quad \sum_{i = 1}^m \|W_i\|_2^2\cdot h\left(\frac{\left\|\Sigma \txt{ diag}\left(\Lambda\right)V_i\right\|_2}{\|\Sigma V_i\|_2}\right)
\]
where $W = U\Sigma V^{\top}$ with $U \in \mathbb{R}^{d \times d}, \Sigma \in \mathbb{R}^{d \times d}$ is diagonal, and $V \in \mathbb{R}^{m \times d}$, with $U^{\top}U = I_{d \times d}, V^{\top}V = I_{d \times d}$ and $V_i \in \mathbb{R}^d$ is the $i^{th}$ column of $V^{\top}$, and where $h$ is defined in Definition~\ref{def:hfunc}.
Then, the optimal rank-$r$ matrix $Y^* \in \mathbb{R}^{d \times m}$ minimizing the initial objective ~\eqref{eq:population_objective} is given by 
$
Y^* = U \txt{ gnorm}\left(\txt{ diag}\left(\Lambda^*\right)\Sigma V^{\top}\right),
$
where $
\txt{gnorm}(A)_i := A_i \cdot \frac{\sqrt{h}(\|A_i\|_2)}{\|A_i\|_2}
$
for the $i^\txt{th}$ column of matrix $A$.
\end{theorem}
\begin{proof}
First, we expand
\begin{align}
\begin{split}
    &\mathbb{E}_{x \sim \mathcal{N}(0, I)}\left[\left\|\sigma(x^{\top}Y) - \sigma(x^{\top}W)\right\|_2^2\right] 
    \\
    &= \mathbb{E}_{x \sim \mathcal{N}(0, I)}\left[\left\|\sigma(x^{\top}Y)\right\|_2^2\right] + \mathbb{E}_{x \sim \mathcal{N}(0, I)}\left[\left\|\sigma(x^{\top}W)\right\|_2^2\right] 
    \\
    &- 2\mathbb{E}_{x \sim \mathcal{N}(0, I)}\left[\langle\sigma(x^{\top}Y), \sigma(x^{\top}W) \rangle\right]
    \\
    &= C + \frac{1}{2}\left\|Y\right\|_F^2 - 2\mathbb{E}_{x \sim \mathcal{N}(0, I)}\left[\langle\sigma(x^{\top}Y), \sigma(x^{\top}W) \rangle\right]
    \\
    &= C + \frac{1}{2}\left\|Y\right\|_F^2 - \sum_{i = 1}^m \|W_i\|_2\|Y_i\|_2\cdot  \sqrt{h}\left(\frac{Y_i^{\top}W_i}{\|W_i\|_2\|Y_i\|_2}\right)
    \\
    &= C - \sum_{i = 1}^m \left[ \left(\|W_i\|_2\sqrt{h}(\rho_i)\right)\beta_i - \frac{1}{2}\beta_i^2\right]
\end{split}
\end{align}
where $C = \frac{1}{2}\|W\|_F^2$ is a constant independent of choice of $Y$, and letting $\rho_i = \frac{Y_i^{\top}W_i}{\|Y_i\|_2\|W_i\|_2}$ and $\beta_i = \|Y_i\|_2$, where $Y_i$ and $W_i$ are column $i$ of $Y, W$ respectively. In the above display, we used Lemma~\ref{relu_norm} and Lemma~\ref{lem:hermite_analysis_correlation}.

Now, we can re-write the minimization problem as a maximization problem: 
\[
\max_{\rho, \beta: Y \in \mathbb{R}^{d \times m} \textnormal{ is rank }r} \sum_{i = 1}^m \left[ \left(\|W_i\|_2\sqrt{h}(\rho_i)\right)\beta_i - \frac{1}{2}\beta_i^2\right]
\]
Note we can write this as an optimization problem over just $\rho \in \mathbb{R}^m$, since the choice of norm $\beta_i = \|Y_i\|_2$ is independent of the value of $\rho_i$. Since the objective as a function of $\beta_i$ is separable and concave in each term of the sum, we can easily solve the maximization problem by setting the derivative to $0$: 
\[
\beta_i^* = \|W_i\|_2\sqrt{h}(\rho_i)
\]
Plugging in this optimal choice of $\beta_i$ for any choice of $\rho$ (and remembering that we must correctly re-normalize later), we get the new objective:

\begin{equation}\label{eq:weighted_relu_svd}
\max_{\substack{\rho: Y \in \mathbb{R}^{d \times m} \textnormal{ is rank }r \\ \|Y_i\|_2 = \|W_i\|_2 \sqrt{h}(\rho_i)}} \frac{1}{2}\sum_{i = 1}^m \|W_i\|_2^2 \cdot h(\rho_i) 
\end{equation}

Now, all that remains is to parameterize low-rank $Y$ as a function of $\rho$.
Let us write $W = U\Sigma V^{\top}$ and $Y = D\Lambda B^{\top}$ via singular value decomposition, understanding that $\Sigma \in \mathbb{R}^{d \times d}$ is a full-rank diagonal matrix and $\Lambda \in \mathbb{R}^{d \times d}$ is a rank-$r$ diagonal matrix (with only $r$ non-zero terms on the diagonal), and $U^{\top}U = I, V^{\top}V = I, D^{\top}D = I, B^{\top}B = I$ are orthonormal matrices. Now we can see how to choose $D, B, $ and $\Lambda$ to maximize the objective. We have
\[
\rho_i = \frac{B_i^{\top}\Lambda D^{\top} U \Sigma V_i}{\|B_i^{\top}\Lambda\|_2 \|W_i\|_2}
\]
where $B_i \in \mathbb{R}^d$ is the $i^{th}$ column of $B^{\top}$ and $V_i \in \mathbb{R}^d$ is the $i^{th}$ column of $V^{\top}$
, noting that $\|B_i^{\top}\Lambda D^{\top}\|_2 = \|B_i^{\top}\Lambda\|_2$ since $D$ is orthonormal.

Since $h$ is convex and monotone-increasing, we want to maximize the value of $\rho_i$ subject to the low-rank constraint whenever it is the case that the parameter choices made for $\rho_i$ do not affect the other $\rho_i$ (since overall we want to maximize $\sum_{i = 1}^m \|W_i\|_2^2\cdot h(\rho_i)$). We proceed by identifying the optimal choice of parameters for each of $B, D,$ and $\Lambda$. For any choice of $B, \Lambda$, the optimal choice of $D$ is $U$ since $D$ is constrained to be orthonormal. Now fix any choice of diagonal rank-$r$ $\Lambda^*$. We wish to maximize the following over $B_i$:
\[
\max_{\|\Lambda^*B_i\|_2 = 1} \frac{B_i^{\top}\Lambda^*\Sigma V_i}{\|B_i^{\top}\Lambda^*\|_2\|W_i\|_2}
\]
Choosing $w = \Lambda^*B_i$ (and now working with the versions of $B_i, V_i \in \mathbb{R}^r$, where the $r$ dimensions correspond to the dimensions which are non-zero on the diagonal for $\Lambda^* \in \mathbb{R}^{d \times d}$)\footnote{This minor step is important since it defines the inverse ${\Lambda^*}^{-1} \in \mathbb{R}^{r \times r}$. We switch between the $r$-dimensional version and the $d$-dimensional version freely, where the version being used depends on context.}, we get the problem 
\[
\max_{\|w\|_2 = 1} \frac{w^{\top}\Sigma V_i}{\|w\|_2\cdot\|W_i\|_2}
\]
for which the solution is $w = \Sigma V_i$ (by Cauchy-Schwarz).
Then, we get
\[
B_i = {\Lambda^*}^{-1}\Sigma V_i
\]
and plugging it in, we get that with this choice of $B_i$, the correlations are
\begin{align}
\begin{split}
    \rho_i &= \frac{B_i^{\top}\Lambda^*\Sigma V_i}{\|B_i^{\top}\Lambda^*\|_2\|W_i\|_2} = \frac{V_i^{\top}\Sigma_{r \times r}^2 V_i}{\|\Sigma_{r \times r}V_i\|_2\|W_i\|_2} 
    \\
    &= \frac{\|\Sigma_{r \times r}V_i\|_2}{\|W_i\|_2} = \frac{\|\Sigma_{d \times d}\Lambda^*V_i\|_2}{\|\Sigma_{d\times d}V_i\|_2}
\end{split}
\end{align}
since we note $\|W_i\|_2 = \|U\Sigma V_i\|_2 = \|\Sigma V_i\|_2$ and
where we now restrict $\Lambda^* \in \mathbb{R}^{d \times d}$ to be a diagonal matrix with $r$ ones on the diagonal and the rest of the entries zero. At the end we also interchanged between $V_i \in \mathbb{R}^r$ (with $r$ coordinates chosen by $\Lambda^*$) and $V_i \in \mathbb{R}^d$.

\end{proof}

\begin{lemma}[ReLU Decomposition]
Let ReLU be defined by $\sigma(x) = \max(0, x)$.
\begin{align}
    \sigma(x) = \frac{x + |x|}{2}
\end{align}
\end{lemma}
\begin{proof}
Observe that if $x \leq 0$, $\sigma(x) = 0$. Otherwise, it equals $x$.
\end{proof}

\begin{lemma}[ReLU Norm]\label{relu_norm}
\begin{align}
    \mathbb{E}_{x \sim \mathcal{N}(0, I)}\left[\left\|\sigma(x^{\top}Y)\right\|_2^2\right] &= \frac{1}{2}\left\|Y\right\|_F^2
\end{align}
\end{lemma}
\begin{proof}
We have 
\begin{align}
\begin{split}
    \left\|\sigma(Y^{\top}x)\right\|_2^2 &= \sum_{i = 1}^m \sigma(Y_i^{\top}x)^2
    \\
    &= \frac{1}{4}\sum_{i = 1}^m \left(Y_i^{\top}x + |Y_i^{\top}x|\right)^2
    \\
    &= \frac{1}{4}\sum_{i = 1}^m 2\left(Y_i^{\top}x\right)^2 + 2\left(Y_i^{\top}x\right)|Y_i^{\top}x|
    \\
    &= \frac{1}{2}\left\|Y^{\top}x\right\|_2^2 + \frac{1}{2}\sum_{i = 1}^m \left(Y_i^{\top}x\right)^2 \textnormal{sgn}(Y_i^{\top}x)
\end{split}
\end{align}
Now we take expectation over $x \sim \mathcal{N}(0, I)$.
\begin{align}
\begin{split}
    \mathbb{E}_{x \sim \mathcal{N}(0, I)}\left[\left\|Y^{\top}x\right\|_2^2\right] &= \mathbb{E}_{x \sim \mathcal{N}(0, I)}\left[\textnormal{Tr}\left(x^{\top}YY^{\top}x\right)\right] 
    \\
    &= \textnormal{Tr}\left(Y^{\top}\mathbb{E}_{x \sim \mathcal{N}(0, I)}\left[xx^{\top}\right]Y\right)
    \\
    &= \textnormal{Tr}\left(Y^{\top}Y\right)
    \\
    &= \left\|Y\right\|_F^2
\end{split}
\end{align}
For the second term, apply linearity of expectation and condition on the events that $Y_i^{\top}x > 0$ and $Y_i^{\top}x < 0$. Since $Y_i$ is fixed and $x$ is an isotropic Gaussian (which is spherically symmetric), the probability that $Y_i^{\top}x > 0$ is equal to the probability that $Y_i^{\top}x < 0$. Thus, letting $q_i = \mathbb{P}\left(Y_i^{\top}x > 0\right)$,
\begin{align}
\begin{split}
    q_i\cdot &\sum_{i = 1}^m \mathbb{E}_{x \sim \mathcal{N}(0, I)}\left[\left(Y_i^{\top}x\right)^2\textnormal{sgn}(Y_i^{\top}x) | Y_i^{\top}x > 0\right]
    \\ +  (1 - q_i)\cdot &\sum_{i = 1}^m \mathbb{E}_{x \sim \mathcal{N}(0, I)}\left[\left(Y_i^{\top}x\right)^2\textnormal{sgn}(Y_i^{\top}x) | Y_i^{\top}x < 0\right] 
    \\
    &= 0
\end{split}
\end{align}
and we get the desired result.
\end{proof}

\begin{lemma}[Reduction to $1$-Dimensional Correlation]
Consider $W, Y \in \mathbb{R}^{d \times m}$, and ReLU nonlinearity $\sigma$, which is applied elementwise. Then,
\begin{align}
&\mathbb{E}_{x \sim \mathcal{N}(0, I)}\left[\langle\sigma(x^{\top}Y), \sigma(x^{\top}W) \rangle\right] 
\\
&= \sum_{i = 1}^m \|W_i\|_2\|Y_i\|_2 \cdot \E{g_i^1, g_i^2}{\sigma(g_i^1)\sigma(g_i^2)}
\end{align}
where $g_i^1, g_i^2$ are univariate standard Gaussians with correlation $\E{g_i^1, g_i^2}{{g_i^1}^{\top}g_i^2} = \rho_i$ and $\rho_i = \frac{W_i^{\top}Y_i}{\|W_i\|_2\|Y_i\|_2} \in [0, 1]$. $W_i, Y_i$ are the columns of $W, Y$ respectively.
\end{lemma}
\begin{proof}
First apply linearity of expectation to get a sum over $m$ correlations corresponding to column vectors $Y_i, W_i$. Then, using the positive homogeneous property of ReLU ($\sigma(c\cdot x) = c\cdot \sigma(x)$ for $c \geq 0$), we can normalize $Y_i, W_i$ and pull out their $\ell_2$ norms. Then using joint Gaussinity, we can replace $Y_i^{\top}x$ and $W_i^{\top}x$ with $g_i^1, g_i^2$. Finally, observe
\begin{align}
\begin{split}
    \E{x \sim \mathcal{N}(0, I)}{\txt{Tr}\left(Y_i^{\top}xx^{\top}W_i\right)} \cdot \frac{1}{\|Y_i\|_2\|W_i\|_2} &= \frac{\txt{Tr}\left(Y_i^{\top}I W_i\right)}{\|Y_i\|_2\|W_i\|_2} 
    \\ &= \frac{Y_i^{\top}W_i}{\|Y_i\|_2\|W_i\|_2}
\end{split}
\end{align}

\end{proof}

\subsubsection{Lower Bounding the Gap}

\begin{lemma}[Relationship between SVD and ReLU SVD]
Suppose $W = U\Sigma V^{\top} \in \mathbb{R}^{d \times m}$.
Consider the solution for rank $r$ ReLU SVD (given by $Y^*$) as described in Theorem~\ref{thm:relu_svd}:
\[
Y^* = U \txt{ gnorm}\left(\txt{ diag}\left(\Lambda^*\right)\Sigma V^{\top}\right)
\]
where the $\txt{gnorm}(\cdot)$ operation is defined by
\[
\txt{gnorm}(A)_i := A_i \cdot \frac{\sqrt{h}(\|A_i\|_2)}{\|A_i\|_2}
\]
for column $i$ of matrix $A$.
If $h(\rho)$ is replaced with $\rho^2$, and we always choose $\Lambda^*$ to correspond to the top $r$ singular values of $\Sigma$, $Y^*$ is the standard SVD solution.
\end{lemma}
\begin{proof}
By direct evaluation of the formula of $Y^*$ while replacing $h(\rho)$ with $\rho^2$ -- note that the terms $\|W_i\|_2^2$ get canceled in this special case.
\end{proof}

\begin{theorem}[Lower Bound on Suboptimality of SVD]
Recall the objective $\mathcal{R}(Y)$
from Problem~\ref{prob:learning_prob_statement},
where we require that $Y \in \mathbb{R}^{d \times m}$ is a rank-$r$ matrix.
Let $W \in \mathbb{R}^{d \times m}$. Define $\rho_{\sigma}^* \in \mathbb{R}^m$ as the correlations $\|\Sigma \txt{ diag}\left(\Lambda^*\right) V_i\|_2 / \|\Sigma V_i\|_2$, where $\Lambda^* \in \{0, 1\}^d$ is the optimum of the optimization problem posed in Theorem~\ref{thm:relu_svd}. As shorthand, denote $Y(\rho)$ as the associated low-rank matrix for correlation vector $\rho \in \mathbb{R}^m$ (computed as described in Theorem~\ref{thm:relu_svd}). Denote $\rho_{\txt{SVD}}^*$ to be the optimal correlations in the case where we pick $\Lambda^*$ to correspond to the top $r$ singular values, as in SVD.
Then, we have the following lower bound for the suboptimality of the SVD solution $Y_{\txt{SVD}}$:
\begin{align}
\begin{split}
     \frac{1}{d}\left(\mathcal{R}(Y_{\txt{SVD}})) - \mathcal{R}(Y(\rho_{\sigma}^*))\right) \geq \frac{1}{2d}\|w\odot \sqrt{h(\rho_{\txt{SVD}}^*)} - \rho_{\txt{SVD}}^*\|_2^2
\end{split}
\end{align}
where $h$ is defined in Definition~\ref{def:hfunc}, where we normalize by $d$ to get an average over the entry-wise error of the approximated output, where $w = \begin{bmatrix} \|W_1\|_1 & \cdots * \|W_m\|_2 \end{bmatrix}$ is a vector of column norms of $W$, and where $\odot$ is the elmeentwise product.
\end{theorem}
\begin{proof}
First consider that the value of the ReLU SVD objective as a function of arbitrary $\rho$ and $\beta_i = \|Y_i\|_2$ for column $i$ of $Y$ is
\[
\frac{\|W\|_F^2}{2} - \sum_{i = 1}^m \left[ \|W_i\|_2 \sqrt{h}(\rho_i)\beta_i - \frac{1}{2}\beta_i^2\right]
\]
as computed in Theorem~\ref{thm:relu_svd}. We know that choosing $\Lambda^*$ to correspond to the top $r$ singular values is not necessarily optimal, and so if we bound the difference between the SVD solution and the choice of $\rho_{\txt{SVD}}^*$ with correct scaling, we are also lower bounding the difference between the SVD solution and the optimal choice of $\rho_{\sigma}^*$ (and correct scaling). If we choose $\Lambda^*$ to correspond to the top $r$ singular values (sub-optimally) and then set $\beta_i = \|W_i\|_2\cdot \sqrt{h}(\rho_{\txt{SVD}}^*(i))$ optimally, we get that the value of the objective is 
\[
\frac{\|W\|_F^2}{2} - \frac{1}{2}\sum_{i = 1}^m \|W_i\|_2^2\cdot h(\rho_{\txt{SVD}}^*(i))
\]
On the other hand, if we do the SVD solution, we can note that since we pick the same $\Lambda^*$, the SVD solution only differs in that we plug in $\beta_i = \rho_{\txt{SVD}}^*(i)$ to the objective instead (Lemma~\ref{lemma:svd_relusvd_relation}):
Thus, the resulting value of that objective is
\[
\frac{\|W\|_F^2}{2} - \sum_{i = 1}^m \left[ \|W_i\|_2\cdot \sqrt{h}(\rho_{\txt{SVD}}^*(i))\rho_{\txt{SVD}}^*(i) - \frac{1}{2}{\rho_{\txt{SVD}}^*(i)}^2\right]
\]
Therefore, computing the absolute difference to get the suboptimality gap, we get
\begin{align}
\begin{split}
    &\sum_{i = 1}^m \|W_i\|_2\cdot \sqrt{h}(\rho_{\txt{SVD}}^*(i))\rho_{\txt{SVD}}^*(i) - \frac{1}{2}{\rho_{\txt{SVD}}^*(i)}^2 
    \\
    &+ \sum_{i = 1}^m \frac{1}{2}\|W_i\|_2^2 \cdot h(\rho_{\txt{SVD}}^*(i))
    \\
    &= \frac{1}{2}\sum_{i = 1}^m\left(\|W_i\|_2\cdot \sqrt{h}(\rho_{\txt{SVD}}^*(i)) - \rho_{\txt{SVD}}^*(i)\right)^2
\end{split}
\end{align}
which after normalizing by $d$ proves the result.
\end{proof}

Using the lower bound in Theorem~\ref{thm:subopt_lowerbound}, we can now characterize the conditions on the full-rank matrix $W \in \R^{d \times m}$ where the optimum of $\mathcal{R}(Y)$ from Problem~\ref{prob:learning_prob_statement} and the SVD solution are significantly different.

\begin{corollary}\label{cor:small_rho_larger_gap}
Consider $W \in \R^{d \times m}$ and the problem of finding the best nonlinear approximation low-rank $Y \in \R^{d \times m}$, as described in Theorem~\ref{thm:relu_svd}. Then, cases where more correlation terms $\rho_{\txt{SVD}}^*(i)$ are smaller result in the SVD solution having larger sub-optimality gaps (measured with respect to $\mathcal{R}(Y)$ defined in Problem~\ref{prob:learning_prob_statement}). 
\end{corollary}
\begin{proof}
From the proofs of Theorem~\ref{thm:relu_svd} and Theorem~\ref{thm:subopt_lowerbound}, we only need to consider the difference between $\sqrt{h}(\rho_{\txt{SVD}}^*(i))$ and $\rho_{\txt{SVD}}^*(i)$.
Note that the gap between $\sqrt{h}(\rho_{\txt{SVD}}^*(i))$ and $\rho_{\txt{SVD}}^*(i)$ increases as $\rho_{\txt{SVD}}^*(i)$ decreases (since $\sqrt{h}(\rho_{\txt{SVD}}^*(i))$ is a convex and monotone increasing upper bound to $\rho_{\txt{SVD}}^*(i)$ which converges at $\rho_{\txt{SVD}}^*(i) = 1$, see Lemma~\ref{lem:h_convex_monotone} and Corollary~\ref{def:hfunc}).
\end{proof}

\begin{remark}[Orthogonality Intuition for Corollary~\ref{cor:small_rho_larger_gap}]\label{rem:orthogonality_intuition}
Corollary~\ref{cor:small_rho_larger_gap} roughly translates to a statement about the approximate orthogonality of the columns of $W$ -- if the columns of $W$ are all mostly orthogonal, it is hard to achieve large correlations $\rho_i = \langle \frac{W_i}{\|W_i\|_2}, \frac{Y_i}{\|Y_i\|_2} \rangle$ with a low-rank $Y \in \R^{d \times m}$ since we can only select $Y_i$ which span a certain subspace -- if the columns are approximately orthogonal, they do not live in a degenerate subspace and if for instance they are chosen uniformly randomly on the sphere, they will typically be near orthogonal to the vectors of the low-rank subspace spanned by the columns of $Y$. Therefore, since the $\rho_i$ will be typically small in this setting, by Corollary~\ref{cor:small_rho_larger_gap}, there will be a larger sub-optimality gap for the SVD.
\end{remark}

When $\max_{i} \rho_{\txt{SVD}}^*(i) < 1$ (larger $\rho_{\txt{SVD}}^*(i)$ correspond to smaller gaps), we can prove a stronger result:

\begin{corollary}\label{cor:dim_increase_rank_decrease}
Suppose the columns of $W \in \R^{d \times m}$ satisfy $\|W_i\|_2 = 1$ for all $i\in [m]$. Then, the suboptimality of SVD is lower bounded in terms of the growth rates of $m, d$. If $d = \Theta(n), m = \Omega(n^{1 + \delta})$ for any $\delta > 0$, and we have that $\max_{i \in [m]} \rho_{\txt{SVD}}^*(i) < \rho_{\max} < 1$, then the sub-optimality gap grows as $d$ increases. Furthermore, the sub-optimality gap is monotone non-decreasing as $r$ decreases.
\end{corollary}
\begin{proof}
Using Theorem~\ref{thm:subopt_lowerbound} and the fact that $\|W_i\|_2 = 1$ for all $i \in [m]$, we have
\begin{align}
\begin{split}
    &\frac{1}{2d}\left\|\sqrt{h}(\rho_{\txt{SVD}}^*) - \rho_{\txt{SVD}}^*\right\|_2^2 
    \\
    &= \frac{1}{2d}\sum_{i = 1}^m \left(\sqrt{h}(\rho_{\txt{SVD}}^*(i)) - \rho_{\txt{SVD}}^*(i)\right)^2 
    \\
    &\geq \frac{m}{2d}\cdot \left(\sqrt{h}(\rho_{\max}) - \rho_{\max}\right)^2
    \\
    &\geq \frac{c_1\cdot n^{1 + \delta}}{c_2 \cdot n}\cdot \left(\sqrt{h}(\rho_{\max}) - \rho_{\max}\right)^2
    \\
    &\geq \Omega\left(d^{\delta} \cdot \left(\sqrt{h}(\rho_{\max}) - \rho_{\max}\right)^2\right)
\end{split}
\end{align}
where we used the fact that $\sqrt{h}(\rho_{\txt{SVD}}^*(i)) - \rho_{\txt{SVD}}^*(i)$ decreases as $\rho_{\txt{SVD}}^*(i)$ increases. Then note that the maximum correlation attainable by the low-rank SVD approximation, $\rho_{\max}$, is a monotone non-decreasing function of the rank $r$. This fact follows because by reducing the rank, we only further restrict the choice of subspace which the columns $Y_i \in \R^r$ can live in. Applying this fact, we have that as $r$ decreases, by Corollary~\ref{cor:small_rho_larger_gap}, $\left(\sqrt{h}(\rho_{\max}^*) - \rho_{\max}^*\right)^2$ does not decrease.
\end{proof}

\begin{corollary}[Spherical Weights]
Suppose the columns of $W \in \R^{d \times m}$ are drawn from the uniform distribution over the surface of the unit sphere. Consider target rank $r \leq d$. Assume that $r \gg \Theta(\log(m))$. Then with probability at least constant,
\[
\rho_{\max} \leq \sqrt{\frac{r}{d}}
\]
and thus
\begin{align}
\begin{split}
    &\frac{1}{d}\left(\mathcal{R}(Y_{\txt{SVD}})) - \mathcal{R}(Y(\rho_{\sigma}^*))\right) 
    \\
    &\geq \frac{1}{2d}\left\|\sqrt{h}(\rho_{\txt{SVD}}^*) - \rho_{\txt{SVD}}^*\right\|_2^2 
    \\
    &\geq \frac{m}{2d}\cdot \left(\frac{1}{\pi} - \frac{1}{2} \sqrt{\frac{r}{d}}\right)^2
    \\
    &\geq \Omega\left(d^{\delta}\cdot \left(\frac{1}{\pi} - \frac{1}{2} \sqrt{\frac{r}{d}}\right)^2\right)
\end{split}
\end{align}
assuming that $m = d^{1 + \delta}$ for some $\delta > 0$
and that $\sqrt{\frac{r}{d}} < \frac{2}{\pi}$.

More generally, we can plug in the formula for $\sqrt{h}(\rho)$ given in Definition~\ref{def:arccos_kernel} without restricting the values of $\frac{r}{d}$.
Therefore, we see that in the case of spherical $W$, the sub-optimality gap increases as either the dimension increases (when width is super-linear in dimension) or as the rank scale decreases.
\end{corollary}
\begin{proof}
The key step in this proof is to upper bound $\rho_{\max}$ with high probability. Then we can apply our lower bound from Corollary~\ref{cor:dim_increase_rank_decrease}. We have 
\[
\rho_{\txt{SVD}}^*(i) = \|\Sigma E_r V_i\|_2 / \|W_i\|_2
\]
where $W = U \Sigma V^{\top}$ with $V_i \in R^d$ being a column of $V^{\top}$ and $E_r = \txt{ diag}\left(\Lambda^*\right)$, where $\Lambda^* \in \{0, 1\}^{d}$ is a binary vector with $r$ non-zero entries selecting the top $r$ singular vectors of $W$. Note $\rho_{\txt{SVD}}^*(i) \leq 1$ since $\|W_i\|_2 = \|\Sigma V_i\|_2$. Thus, we have

\[
\frac{\|\Sigma E_r V_i\|_2}{\|\Sigma V_i\|_2} = \frac{\|E_r \Sigma V_i\|_2}{\|\Sigma V_i\|_2} = \left\|E_r \frac{\Sigma V_i}{\|\Sigma V_i\|_2}\right\|_2
\]
which holds since $\Sigma$ is diagonal.

Now, treating $W$ and its SVD as random variables, note that $W = U\Sigma V^{\top}$, whose columns are uniformly distributed on the surface of the sphere, can be written as a matrix of $m$ i.i.d. Gaussian vectors drawn from $\mathcal{N}\left(0, I_{d \times d}\right)$ with normalized columns, and that $U \in \R^{d \times d}$ is an orthogonal matrix. By the fact that the Gaussian is rotationally symmetric, multiplying by $U^{\top}$ does not change the distribution. Thus we find that $\Sigma V^{\top} \in R^{d \times m}$ is also a matrix of $m$ independent Gaussians distributed according to $\mathcal{N}\left(0, I_{d \times d}\right)$. Note that the specific $r$ non-zero entries chosen does not matter by spherical symmetry (we can multiply by a permutation matrix, which is a rotation matrix, and not affect the distribution). Thus, we can write
\[
\rho_{\max} = \max_{i \in [m]} \left\|E_r \frac{\Sigma V_i}{\|\Sigma V_i\|_2}\right\|_2 = \max_{i \in [m]} \sqrt{\sum_{j = 1}^r \frac{g_{ij}^2}{\|g_i\|_2^2}}
\]
where $g_i \sim \mathcal{N}\left(0, I_{d \times d}\right)$ for $i \in [m]$ are i.i.d. standard Gaussians. Now we apply Talagrand's concentration inequality (as stated in Theorem $4.20$ in \citet{van2016probability}).

We can easily check that the max over the $m$ values of $\rho_{\txt{SVD}}^*(i)$ is a convex, $1$-Lipschitz function, and therefore Talagrand's inequality applies:
\[
\mathbb{P}\left(\left| \rho_{\max} - \E{}{\rho_{\max}}\right| > t \right) \leq C\cdot \exp\left(-c\cdot t^2\right)
\]
where $c, C$ are universal constants (we will abuse the notation $c, C$ throughout to denote universal constants).
All that remains is to bound $\E{}{\rho_{\max}}$ above; then probability that $\rho_{\max}$ deviates above this upper bound by more than $t$ is also still bounded by $C\cdot \exp\left(-c\cdot t^2\right)$.

Thus, we have
\begin{align}
\begin{split}
    \EE{}{\rho_{\max}} &= \EE{}{\max_{i \in [m]} \sqrt{\sum_{j = 1}^r \frac{g_{ij}^2}{\|g_i\|_2^2}}}
\end{split}
\end{align}

Then, since $z_i := \sqrt{\sum_{j = 1}^r \frac{g_{ij}^2}{\|g_i\|_2^2}} \in [0, 1]$ are independent and identically distributed, we get that the distribution for the max over these $m$ variables is given by 
\[
\mathbb{P}\left(\rho_{\max} > t\right) =  1 - \left(1 - \mathbb{P}\left(z > t\right)\right)^m
\]
for $z = z_1$.

Now we seek a better understanding of $\mathbb{P}\left(z > t\right)$ so we can compute an upper bound on $\E{}{\rho_{\max}}$. First we have 
\[
\mathbb{P}\left(\|g_i\|_2 \not\in \sqrt{(1 + \eps)\cdot d}\right) \leq \exp\left(-c\cdot d \cdot \min(\eps, \eps^2)\right)
\]
by standard Gaussian norm concentration (Theorem $3.1.1$ in \citet{vershynin2018high}). 

Then consider the vector $h_i = \begin{bmatrix} g_{i1} & \cdots & g_{ir} \end{bmatrix}$ while also conditioning on the previous bound holding (e.g., conditioning on $\|h_i\|_2^2 \leq (1 + \eps)\cdot d$). Denote this event by $F$. 

After conditioning, these random variables also still satisfy the concentration of Gaussian norm condition since the event $F$ only decreases the probability that $\|h_i\|_2$ does not lie within the band, and we have 
\[
\mathbb{P}\left(\|h_i\|_2 \not\in \sqrt{(1 + \eps)\cdot r} \mid F\right) \leq \exp\left(-c\cdot r \cdot \min(\eps, \eps^2)\right)
\]
Then union bounding over both events gives us that 
\begin{align}
\begin{split}
    &\mathbb{P}\left(\frac{\|h_i\|_2}{\|g_i\|_2} \not\in \left[\sqrt{\frac{1 - \eps}{1 + \eps}\cdot \frac{r}{d}}, \sqrt{\frac{1 + \eps}{1 - \eps}\cdot \frac{r}{d}} \right] \right) \\
    &\leq 2\exp\left(-c\cdot r \cdot \min(\eps, \eps^2)\right)
\end{split}
\end{align}
We can use this fact about $\mathbb{P}(z > t)$ to upper bound the expectation. We have
\begin{align}
\begin{split}
    \E{}{\rho_{\max}} &= \E{}{\max_{i \in [m]} z_i}
    \\
    &= \int_{0}^{\infty} \left(1 - \left(1 - \mathbb{P}\left(z \geq t\right)\right)^m\right) dt
    \\
    &= \int_{0}^{1} \left(1 - \left(1 - \mathbb{P}\left(z \geq t\right)\right)^m\right) dt
    \\
    &\leq \int_{0}^{\sqrt{\frac{1 + \eps}{1 - \eps} \cdot \frac{r}{d}}} 1 \cdot dt 
    \\
    &+ \int_{\sqrt{\frac{1 + \eps}{1 - \eps} \cdot \frac{r}{d}}}^{1} \left(1 - \left(1 - \exp\left(-c\cdot r\cdot \eps^2\right)\right)^m\right) dt
    \\
    &\leq \sqrt{\frac{1 + \eps}{1 - \eps} \cdot \frac{r}{d}} + \left(1 - \sqrt{\frac{1 + \eps}{1 - \eps} \cdot \frac{r}{d}}\right)\cdot \upsilon(m, r)
\end{split}
\end{align}
where we used the fact that $z \in [0, 1]$ and we upper bounded the probabilities by partitioning on some fixed small choice of $\eps < 1$ (say, $0.000001$).
Here, we define
\[
\upsilon(m, r) := 1 - \exp\left(-m\cdot\exp(-\Theta(r\cdot\eps^2))\right)
\]
where we think of $\eps$ as constant. Therefore, $\upsilon(m, r) \in [0, 1]$ and $\upsilon(m, r) \to 1$ if $\log(m)/r \to \infty$, and $\upsilon(m, r) \to 0$ if $\log(m)/r \to 0$. If $\upsilon(m, r) \to 1$, then we see that $\rho_{\max} \to 1$, and if $\upsilon(m, r) \to 0$ (and $\eps \to 0$), $\rho_{\max} \to \sqrt{r/d}$.

Therefore, applying the Talagrand concentration result to the upper bound on $\E{}{\rho_{\max}}$, for $r \gg \log(m)$, we have that with high probability, $\rho_{\max} = \Theta\left(\sqrt{\frac{r}{d}}\right)$.

Now we can use the argument of Corollary~\ref{cor:dim_increase_rank_decrease} and plug in our upper bound on $\rho_{\max}$ to get a lower bound on $\left(\sqrt{h}(\rho_{\max}) - \rho_{\max}\right)^2$ with probability at least $1 - \exp(-\Theta(t^2)) - \exp\left(-\Theta\left(r\cdot \eps^2\right)\right)$:
\begin{align}
\begin{split}
    &\left(\sqrt{h}(\rho_{\max}) - \rho_{\max}\right)^2 
    \\
    &\geq \left(\sqrt{h}\left(t + \sqrt{\frac{1 + \eps}{1 - \eps}\cdot \frac{r}{d}}\right) - \left(t + \sqrt{\frac{1 + \eps}{1 - \eps}\cdot \frac{r}{d}}\right)\right)^2
    \\
    &= \left(\frac{1}{\pi} - \frac{1}{2}\left(t + \sqrt{\frac{1 + \eps}{1 - \eps}\cdot \frac{r}{d}}\right)\right)^2
\end{split}
\end{align}
assuming that $\left(t + \sqrt{\frac{1 + \eps}{1 - \eps}\cdot \frac{r}{d}}\right) < \frac{2}{\pi}$, where we plugged in the approximation
\begin{align}
\begin{split}
    \left(\sqrt{h}(\rho) - \rho\right)^2 &= \left(\frac{\sqrt{1 - \rho^2} + \left(\pi - \txt{arccos}\left(\rho\right)\right)\rho}{\pi} - \rho\right)^2
    \\
    &= \left(\frac{\sqrt{1 - \rho^2}}{\pi} + \left(\frac{\pi - \txt{arccos}(\rho)}{\pi} - 1\right)\rho\right)^2
    \\
    &\geq \left(\frac{1}{\pi}\left(\sqrt{1 - \rho^2} + \rho^2\right) - \frac{\rho}{2}\right)^2
    \\
    &\geq \left(\frac{1}{\pi} - \frac{1}{2}\rho\right)^2
\end{split}
\end{align}
where we used the upper bound $\txt{arccos}(\rho) \leq \frac{\pi}{2} - \rho$ for $\rho \in [0, 1]$. This bound is only valid when $\rho < 2/\pi$ since otherwise the term inside the square is not positive (and it needs to be since $\sqrt{h}(\rho)$ is an upper bound on $\rho$).

More generally, we can just plug in the formula for $\sqrt{h}$ given in Definition~\ref{def:arccos_kernel} to be precise. 

\end{proof}

\begin{figure*}
    \centering
    \includegraphics[scale=0.7]{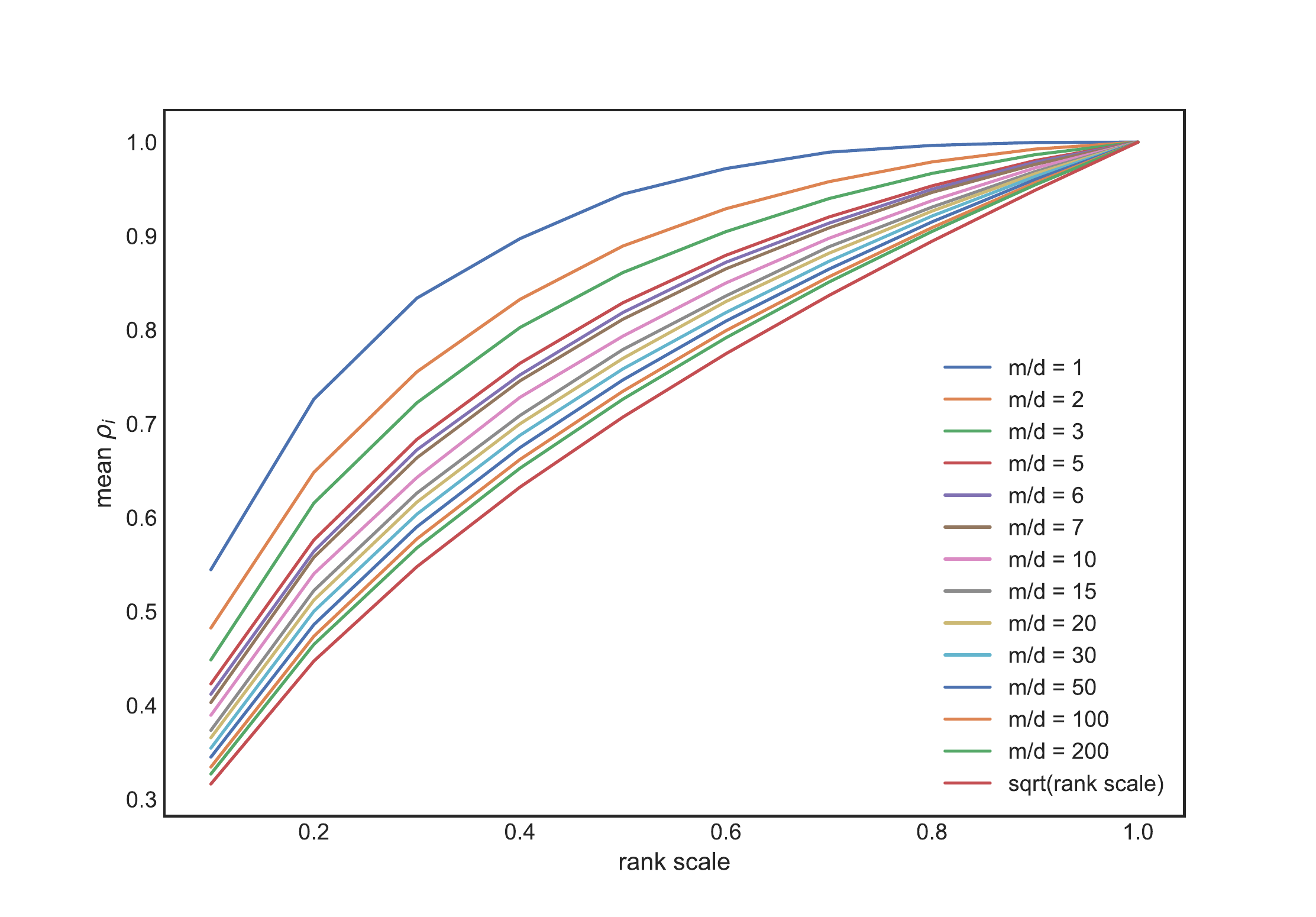}
    \caption{Here we plot the mean values of $\rho_{\txt{SVD}}^*$ for different choices of $m, d, $ and $r$. The standard deviations range between $0.08$ and $0$ (from smallest $r/d = 0.1$ to largest $r/d = 1$). The x-axis is the rank scale $r/d$. We can see that for $m/d$ large enough, the mean corresponds to $\sqrt{\frac{r}{d}}$.}
    \label{fig:typical_rho_vals_gaussian_W}
\end{figure*}

\begin{remark}[What if $\rho_{\max} = 1$?]
It is possible for the upper bound $\rho_{\max} = 1$, even
for low-rank $r$ -- consider a case where most of the columns of $W$
are highly correlated and point in the same direction. It is not difficult to see that one can compute a new column of $W$ which corresponds to one of the columns of $Y_{\txt{SVD}}$ for some choice of rank $r$, and to then add that new column to $W$. The resulting $\hat{W}$ will have an SVD such that one of the columns of $Y$ is identical to this new column of $W$, and thus $\rho_{\max} = 1$. Indeed, in the current proof for Corollary~\ref{cor:spherical_weights_dim_increase_rank_decrease}, we see that if the rank $r$ does not grow sufficiently compared to $m$, the network width, $\rho_{\max} \to 1$ with high probability.

However, this issue is mostly an artefact of our analysis method: In reality, to get an optimal bound for a given $W$, instead of upper bounding $\rho_{\max}$, we should integrate across all the values of $\rho_{\txt{SVD}}^*$ to get our lower bound (e.g. compute the actual $\ell_2$ norm). In the Gaussian case, it turns out that most $\rho_{\txt{SVD}}^*(i) \approx \sqrt{\frac{r}{d}}$ (see Figure~\ref{fig:typical_rho_vals_gaussian_W}), and so to get a rough sense of what the bound looks like, we can simply plug in $\rho_{\txt{SVD}}^*(i) = \sqrt{\frac{r}{d}}$. Note that for constant $r$, as $r$ gets very small, $\sqrt{h}(\rho) - \rho \approx 1/\pi$, and plugging in to Corollary~\ref{cor:dim_increase_rank_decrease}, we get a lower bound on the gap between SVD and the optimum solution to be order $\Omega(d^{\delta})$.
Since this precise control is not essential to our point, we leave out the detailed proof.
\end{remark}

\FloatBarrier

\section{Beyond Low-Rank Approximation}
\label{appendix:beyond_LRA}

The general idea of function approximation at initialization applies to other efficient approximation techniques beyond low-rank deep networks. In particular, we outline how it can be used in the case of general structured matrices via a straightforward extension, as well as in a case where other approximations of a network's weights are baked in (as in \citet{choromanski2020rethinking}).

\begin{definition}[Performer Approximation from \citet{choromanski2020rethinking}]
For $Q, K, V \in \mathbb{R}^{L \times d}$,
let the standard attention layer be defined
\[
\textnormal{Att}_{Q, K}(V) = D^{-1}AV
\]
where we define the \textit{attention matrix} to be
\[
A = \exp\left(QK^T/\sqrt{d}\right)
\]
and the diagonal normalization matrix
\[
D = \textnormal{diag}\left(A \textbf{1}_L\right)
\]
Then, the Performer approximation to the attention is given by a random feature map approximation:
\[
\hat{\textnormal{Att}}_{Q, K}(V) = \hat{D}^{-1}\hat{A}V
\]
where
\[
\hat{A} = \hat{Q}\hat{K}^T; \quad \hat{D} = \textnormal{diag}\left(\hat{Q}\hat{K}^T\textbf{1}_L\right)
\]
and
\[
\hat{Q} = \phi(Q); \quad \hat{K} = \phi(K)
\]
where low-rank (likely randomized) feature map $\phi: \mathbb{R}^d \to \mathbb{R}^r$ is applied to the row vectors. There is a specific family of feature maps that defines the Performer approximation, but our method applies regardless of the choice of $\phi$.
The Performer is more efficient to compute since we can break down the matrix multiplications into an $r \times L$ by $L \times d$ matmul and an $L \times r$ by $r \times d$ matmul, resulting in time complexity $\mathcal{O}(Lrd)$ instead of $\mathcal{O}(L^2 + Ld)$, and with space complexity $\mathcal{O}(Lr + Ld + rd)$.
\end{definition}

For the Performer objective of \citet{choromanski2020rethinking} (and any other approximation of Transformer), layerwise function approximation at initialization yields the following objective:
\begin{definition}[Layerwise Function Approximation at Initialization for Performer]
Suppose we have a neural net with Transformer blocks with substructure 
\[
f_{\theta}(V) = g_{w}(\textnormal{Att}_{Q, K}(V))
\]
defined by $g: \mathbb{R}^{L \times d} \to \mathbb{R}^n$.
The Performer version of the network is instead
\[
\hat{f}_{\theta}(V) = g_{w}(\hat{\textnormal{Att}}_{Q, K}(V))
\]
where $w \in \mathbb{R}^D$ and $\theta = (w, Q, K)$. Then, our procedure is as follows:
\begin{enumerate}
    \item Sample $\theta \sim \mathcal{D}$;
    \item Solve $\min_{\nu \in \mathbb{R}^{D \times 2(L \times d)}} \mathbb{E}_{V_{ij} \sim N(0, 1/\sqrt{L*d})}\left[\left\|f_{\theta}(V) - \hat{f}_{\nu}(V)\right\|_2^2\right]$
    \item Use $\nu^*$ as the initialization.
\end{enumerate}
\end{definition}

As one can see, this general strategy can be applied to any layer approximation method, given an original initialization scheme $\mathcal{D}$ and the ability to optimize the weights to match the original initialization weights. The key element in our approach is taking more of the function into account compared to simply trying to match parameters.

\end{document}